# Robust and Globally Optimal Manhattan Frame Estimation in Near Real Time

Kyungdon Joo *Student Member, IEEE,* Tae-Hyun Oh *Member, IEEE,*
Junsik Kim, *Student Member, IEEE,* In So Kweon, *Member, IEEE*

**Abstract**—Most man-made environments, such as urban and indoor scenes, consist of a set of parallel and orthogonal planar structures. These structures are approximated by the Manhattan world assumption, in which notion can be represented as a Manhattan frame (MF). Given a set of inputs such as surface normals or vanishing points, we pose an MF estimation problem as a consensus set maximization that maximizes the number of inliers over the rotation search space. Conventionally, this problem can be solved by a branch-and-bound framework, which mathematically guarantees global optimality. However, the computational time of the conventional branch-and-bound algorithms is rather far from real-time. In this paper, we propose a novel bound computation method on an efficient measurement domain for MF estimation, *i.e.*, the extended Gaussian image (EGI). By relaxing the original problem, we can compute the bound with a constant complexity, while preserving global optimality. Furthermore, we quantitatively and qualitatively demonstrate the performance of the proposed method for various synthetic and real-world data. We also show the versatility of our approach through three different applications: extension to multiple MF estimation, 3D rotation based video stabilization, and vanishing point estimation (line clustering).

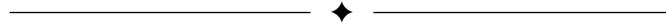

## 1 INTRODUCTION

I<span>N</span> scene understanding, we usually look at the big picture (*i.e.*, a structure of the scene) first and catch details of the scene for a deeper understanding. In case of man-made environments surrounding us, the scenes have structural forms (from the layout of a city to buildings and many indoor objects such as furniture), which can be represented by a set of parallel and orthogonal planes. In the fields of computer vision and robotics, these structures are commonly approximated by Manhattan world (MW) assumption [1]. Under the MW assumption, three orthogonal directions, typically corresponding to X-, Y-, and Z-axes, are used to represent a scene structure, which are referred to as the Manhattan frame (MF) in previous studies [2], [3], [4], [5].[1] Recent studies have proposed a variety of MF estimation methods for scene representation [6], [7], [8], [9]. In addition, MF estimation has been utilized as a key module for various vision applications [2], [10], [11], [12], [13], [14], [15], [16].

In order to ensure the versatile applicability to a broad range, a given MF estimation is required to have two properties: *stability* and *efficiency*. For stability, MF estimation needs to be robust against noise and outliers, and be accurate enough, since it could affect the overall performance of the subsequent applications. Also, it has to be insensitive to initialization, *i.e.*, guarantee of a global optimum. As for efficiency, the computational complexity of the MF estimation has to remain reasonable. Even when its stability is guaranteed, an MF estimation with a high-order complexity is undesirable for time-critical applications, *e.g.*, SLAM and navigation.

In this research, we propose a robust and fast MF estimation approach that guarantees globally optimal solution satisfying both stability and efficiency. We pose the MF estimation problem as a consensus set maximization, and solve it through a branch-and-bound (BnB) framework [17]. Typically, bound computation is one of the main factors that disturbs the computational efficiency of the BnB framework [18]. To resolve this, we suggest relaxing the original problem and define a new bound that can be efficiently computed on a 2D domain (*i.e.*, extended Gaussian image, EGI [19]). This allows the bounds to be computed using a few simple arithmetic operations with constant complexity, while still preserving the global optimality and the convergence property in our BnB framework. The proposed framework is illustrated in Fig. 1. Our method is quantitatively and qualitatively validated with synthetic and real data. We also demonstrate the flexibility of our method in three applications: multiple MF estimation (*a.k.a.*, a mixture of Manhattan frames, MMFs) of a scene, 3D rotation based video stabilization, and vanishing point estimation (line clustering).

This paper extends our previous work published in [20]. We dedicate our effort to further analyze the behavior of the proposed algorithm in both mathematical and empirical perspectives. Specifically, we introduce several relationships essential for deriving the final results (Proposition 2), and provide detailed elucidation and proofs for each lemma and proposition that support theoretical guarantees of the proposed approach. Also, we show an empirical convergence rate of our method. In the experiment, we quantitatively compare with a few more recent MF estimation methods with both simulation and real data including sequential data. Moreover, we extend the application section (Sec. 8.3) in

---

- *K. Joo, J. Kim, and I. S. Kweon are with the School of Electrical and Computer Engineering, KAIST, Daejeon, Republic of Korea.
  T.-H. Oh is with the Computer Science and Artificial Intelligence Laboratory, Massachusetts Institute of Technology, Cambridge, USA.
  E-mail: {kdjoo369, thoh.mit.edu, mibastro}@gmail.com, iskweon@kaist.ac.kr*

1. In the previous studies, they limited the definition of the MF in surface normal context. In our work, we do not limit modality of the MF.

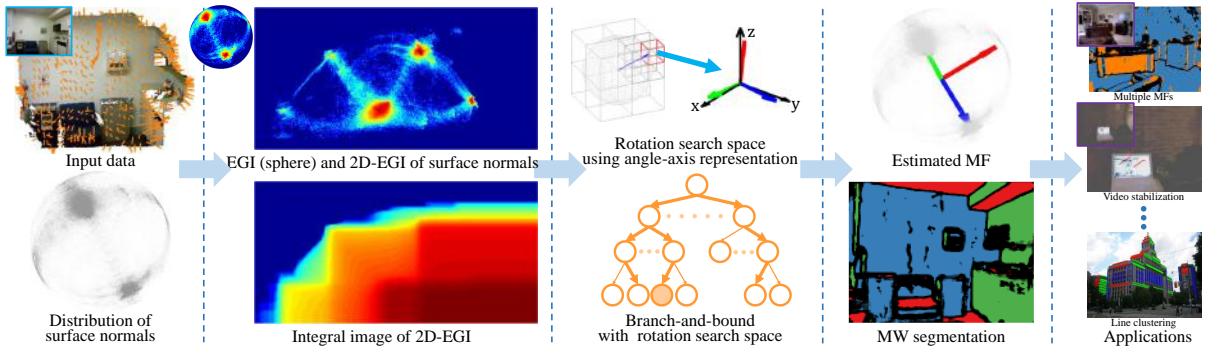

Fig. 1: Overview of the proposed MF estimation approach. *First column*: An example of input data and distribution of surface normals of a scene from the NYUv2 dataset [10]. Not limited to this, VP hypothesis from line pairs can also be fed as input. *Second column*: Its surface normal histogram, *i.e.*, EGI, the 2D-EGI, and its integral image to efficiently calculate bounds. *Third column*: Illustration of the efficient bound based BnB framework using rotation search space. *Fourth column*: The estimated globally optimal MF. *Last column*: Three different applications: Multiple MFs, 3D rotation based video stabilization, and line clustering (vanishing point estimation).

several aspects. We conduct a large-scale experiment with 81 million measurements for MMF estimation, and the quality of video stabilization is improved. We newly provide a vanishing point estimation (line clustering). These demonstrate the versatility of the proposed approach in terms of scalability and stability. The implementation of this work is available online.[2] In summary, the contributions of this work are as follows:

- We propose a branch-and-bound based, fast MF estimation. Our approach can process around $300,000$ measurements in near real-time, and 81 million measurements in near 10 s.
- We relax the problem and present a new and efficient bound computation with constant-time complexity, while guaranteeing a globally optimal solution. Note that the proposed method is inherently robust by definition of consensus set maximization [21], [22].
- Our method has been validated through systematic experiments and real-world data. To show extensibility of our method, we present multiple MF estimations, video stabilization, and line clustering as potential applications.

The remaining parts of the paper are organized as follows. The next section is dedicated to the related work of the MF estimation. In Sec. 3, we formally define the MF estimation problem as a cardinality maximization problem. Based on this formulation, we elucidate a BnB approach most related to our goal through Sec. 4, to clarify its limitations and differences with our proposed method presented in Sec. 5. Then, we lead to the analysis of the proposed method in Sec. 7. Owing to space constraints, all the detailed proofs up to Sec. 7 are placed in the supplementary material. Also, the implementation details are described in Sec. 6. Then, we demonstrate the superiority and versatility of the proposed method with synthetic and real experiments and various applications in Sec. 8. We finally discuss the remaining issues and conclude this paper in Sec. 9. To be concise, we summarize frequently used notations in Table 1.

## 2 RELATED WORK

MF estimation of structured environments under MW assumption [1] is an essential part in high-level vision tasks *e.g.*, scene understanding [2], [10], [14], layout estimation [8], [23], [24], [25], 3D reconstruction [6], [7], focal length estimation [2], [15], *etc*.

2. http://sites.google.com/site/kdjoocv/cvpr2016_mf

| Notation | Description |
|---|---|
| $\mathcal{N} = \{\mathbf{n}_i\}_{i=1}^N$ | A set of surface normals, where $\mathbf{n}_i \in \mathbb{R}^{3\times 1}$ |
| $\mathcal{E} = \{\mathbf{e}_j\}_{j=1}^6$ | A set of six canonical vectors |
| $\mathbf{R} \in SO(3)$ | A $3 \times 3$ rotation matrix |
| $\boldsymbol{\beta} \in \mathbb{R}^{3\times 1}$ | Angle-axis parameterization of rotation matrix |
| $B_\pi$ | A ball of radius $\pi$ |
| $D_{init}$ | An initial cube that tightly encloses the ball $B_\pi$ |
| $c(\cdot)$ | A mapping from a cube to a rotation corresponding to the center of the cube |
| $\mathcal{X}_{\mathbf{L}}, \mathcal{X}_{\mathbf{U}}$ | The boundary point sets of lower and upper inlier regions on the unit sphere manifold |
| $\tilde{\mathcal{X}}_{\mathbf{L}}, \tilde{\mathcal{X}}_{\mathbf{U}}$ | The transferred boundary sets of $\mathcal{X}_{\mathbf{L}}, \mathcal{X}_{\mathbf{U}}$ on 2D-EGI |
| $\mathcal{R}_{\mathbf{L}}, \mathcal{R}_{\mathbf{U}}$ | The rectangular boundary sets tightly enclosing $\tilde{\mathcal{X}}_{\mathbf{L}}, \tilde{\mathcal{X}}_{\mathbf{U}}$ on 2D-EGI |
| $\texttt{EGI} \in \mathbb{R}_+^{m\times n}$ | 2D-EGI representation of the spherical coordinate |
| $s$ | 2D-EGI resolution parameter |
| $\angle(\cdot,\cdot)$ | Angle distance between two vectors on vector space |
| $\phi(\cdot,\cdot)$ | Angle distance along the elevation axis of 2D-EGI |
| $\theta(\cdot,\cdot)$ | Angle distance along the azimuth axis of 2D-EGI |
| $\tau$ | The inlier threshold for the original problem |
| $\tau_{\text{el}}, \tau_{\text{az}}$ | The inlier thresholds for elevation and azimuth axis |
| $L_B, U_B$ | Lower/upper bound for the original problem in Eq. (2) |
| $L_R, U_R$ | Lower/upper bound for the proposed relaxed problem in Eq. (5) |

TABLE 1: Summary of notations.

Depending on the type of input, understanding man-made scene structures (MF estimation) can be boiled down to two methods: estimating orthogonal vanishing points (VPs) in the image domain or estimating three dominant orthogonal directions (*i.e.*, rotation matrix) on a 3D domain, such as depth or 3D point clouds.

Image-based approaches fundamentally rely on the perspective cues of the camera (*i.e.*, a geometric relation between VPs extracted from line segments and Manhattan structure) [6], [8], [12], [26]. There are some works that exploit additional information like structured learning [8], [25], [27] for robustness. We refer the reader to Ghanem *et al*. [3] for a more detailed review about image-based methods. On the other hand, recent studies have focused on accurately estimating dominant orthogonal directions of the scene on 3D using popular Kinect sensor [10], [14], [28]. As long as the related approaches are designed to reveal the Manhattan structure of a scene, we will refer to them as MF estimation approaches for the sake of clarity.

With 3D information, Silberman *et al*. [10] generate all possible MF hypotheses of a scene from both VPs of the image and surface normals. Then, all the hypotheses are exhaustively scored by the number of measurements that support an MF, and then the best one is taken. Note that even though the method exhaustively searches all hypotheses, it does not mean that a whole solution space is searched. For a room layout of indoor scene (*i.e.*, MF), Taylor *et al*. [28] estimate a gravity vector, which corresponds to the floor normal of the ground plane. This approach is based on physical assumptions that the vertical axis should be aligned with the gravity vector and large planar regions of the ground plane should be placed near the bottom of the image. Then, the other orthogonal normal vectors are sequentially estimated from the wall plane segments. Depending on the task, occasionally, only the gravity vector is required. Zhang et al. [29] estimate a floor normal by RANSAC [21] on a set of 3D points whose normals are close to the up-direction. Similar to Taylor et al., Gupta et al. [14] initialize a gravity vector by the y-axis of an RGB-D image and refine it using a simple optimization process. Ghanem et al. [3] have recently proposed a non-convex MF estimation exploiting the inherent sparsity of the membership of each measured normal vector. Unfortunately, all the algorithms mentioned above are sub-optimal.

As the importance of the MF itself has grown, high stability and efficiency of MF estimation are desirable for general purposes. To guarantee stability, Bazin et al. [22], [30] propose an orthogonal VP estimation based on a BnB framework, which guarantees the globally optimal solution. Bazin et al. [30] present an interval analysis-based BnB framework. This strategy is improved in [22] by exploiting the rotation search space [31]. However, BnB frameworks are usually too slow for real-time applications [18]. To alleviate this limitation, Parra et al. [18] propose a fast BnB rotation search method that uses an efficient bound function computation, which can register up to 1000 points in 2 s. However, it is still inadequate for real-time applications. To cope with it, Straub et al. [4] propose a GPU-supported MF inference method that operates in real-time, but does not guarantee global optimality. In contrast to the previous studies, the proposed MF estimation method with BnB framework guarantees a globally optimal solution, as well as a near real-time efficiency.

Recently, Straub et al. [2] suggest a new perspective on MF, which they first call an MMFs. It is a more flexible representation based on a motivation that strict MW assumption cannot represent many real-world man-made scenes. Inspired by this work, we also extend our method to multiple MF estimations as an application.

## 3 PROBLEM STATEMENT

The type of input for our problem can either be 3D surface normals (from depth map or 3D point cloud) or VP candidates (from line segments). For simplicity, we only consider surface normal as the input type in the main context, but we present the application with VP, i.e., line clustering in Sec. 8.3, later. Given a set of surface normals $\mathcal{N} = \{\mathbf{n}_i\}_{i=1}^N$, our goal is to estimate an MF of a scene, which consists of three orthogonal directions.

As implied by the orthogonal property of MF, the process of MF estimation is equivalent to estimating a proper rotation matrix $\mathbf{R} \in SO(3)$, which transforms the standard basis of a coordinate to new three orthogonal bases best aligned with dominant surface normals up to sign. Since a direction vector and its flipping vector indicate the same structural support, we incorporate both the bases of a coordinate and its flipping vectors into a set $\mathcal{E} = \{\mathbf{e}_j\}_{j=1}^6$ [3] of six canonical vectors. Then, estimating the optimal rotation matrix $\mathbf{R}^*$ to the number of inliers can be formulated as the following cardinality maximization problem:

$$\arg\max_{\mathbf{R} \in SO(3)} \sum_{i=1}^N \sum_{j=1}^6 [\![\angle(\mathbf{n}_i, \mathbf{R}\mathbf{e}_j) \leq \tau]\!], \quad (1)$$

3. i.e., $\mathbf{e}_1 = [1\ 0\ 0]^\top$, $\mathbf{e}_2 = [0\ 1\ 0]^\top$, $\mathbf{e}_3 = [0\ 0\ 1]^\top$, $\mathbf{e}_4 = -\mathbf{e}_1$, $\mathbf{e}_5 = -\mathbf{e}_2$, and $\mathbf{e}_6 = -\mathbf{e}_3$.

where $\angle(\mathbf{a}, \mathbf{b})$ is the angle distance between the vectors $\mathbf{a}$ and $\mathbf{b}$, $\tau$ is the inlier threshold, and $[\![\cdot]\!]$ is the indicator function, which outputs 1 if a statement is true, otherwise 0. Specifically, the problem is to find the optimal rotation on the rotation manifold (i.e., solution search space) by counting the number of inlier normals in the input set (i.e., measurement space). Unfortunately, Eq. (1) is intractable to solve directly in a numerical optimization (known to be NP-hard) [30], [32]. Following the approach of Li [32], auxiliary variables $y_{ij} \in \{0, 1\}$ are introduced for indicating whether the $i$-th surface normal $\mathbf{n}_i$ is an inlier ($y_{ij}=1$) or an outlier ($y_{ij}=0$) for the $j$-th direction vector $\mathbf{R}\mathbf{e}_j$. This leads to an equivalent integer programming problem of Eq. (1) as:

$$\begin{aligned}
\arg\max_{\{y_{ij}\}, \mathbf{R} \in SO(3)} \quad & \sum_{i=1}^N \sum_{j=1}^6 y_{ij} \\
\text{s.t.} \quad & y_{ij} \angle(\mathbf{n}_i, \mathbf{R}\mathbf{e}_j) \leq y_{ij} \tau, \\
& y_{ij} \in \{0, 1\}, \forall i = 1 \cdots N,\ j = \{1 \cdots 6\}.
\end{aligned} \quad (2)$$

Solving Eq. (2) is still challenging due to the non-linear geodesic distance measure, the non-linearity of the rotation manifold parameterization, and two families of unknowns entangled in a non-linear way. Although it constitutes a challenging type of non-convex problem, it can be dealt with the BnB framework described in Sec. 4.

## 4 BRANCH-AND-BOUND WITH ROTATION SEARCH

Branch-and-bound (BnB) is a general global optimization framework [17] consisting of *branching* and *bounding* steps. Its basic idea is to recursively divide a solution space into smaller congruent sub-spaces[4] (i.e., *branching* step) and test each sub-space whether it is possible to contain a global optimal solution (i.e., *bounding* step). Infeasible sub-spaces are excluded from the search space at an early stage so that unnecessary computations are avoided.

The keys of BnB framework are to define an appropriate search space and a bound function. In this section, we review the key parts of BnB for MF estimation and discuss its computational limitation.

### 4.1 Branching Part

The first key part of BnB is to define an appropriate search space, which is a rotation space in our MF estimation problem. We employ the angle-axis parameterization to represent a rotation matrix $\mathbf{R}$, which is formed by a three-dimensional vector $\boldsymbol{\beta}$ in a ball $B_\pi$ of radius $\pi$, whose direction $\boldsymbol{\beta}/\|\boldsymbol{\beta}\|$ and norm $\|\boldsymbol{\beta}\|$ specify the axis and angle of a rotation matrix $\mathbf{R}$ [31]. In the angle-axis parameterization, any rotation can be represented by a point in the ball $B_\pi$ which is equivalent to the search space in this problem. Following convention, let $D_{init}$ be an initial cube that tightly encloses the ball $B_\pi$. The cube representation makes the subdivision operation easy because it is axis-aligned. We divide the rotation search space into smaller congruent sub-spaces by octal subdivision of the cube for branching, as illustrated in the third column of Fig. 1.

Given a sub-space partition, i.e., a cube, the next procedure of BnB is to test whether the sub-space could contain a globally optimal solution and then to exclude it for further solution search if it is classified as an infeasible sub-space to contain the optimal solution, as will be described in the following.

---

4. This is irrelevant to the definition of *subspace* in linear algebra literature. Rather, it represents partitions of a solution space. We abuse this terminology throughout this paper without confusion.

**Algorithm 1** The prototype BnB algorithm for rotation space.
1: Initialize the cube list $\mathcal{L}$ with $D_{init}$ s.t. $B_\pi \subset D_{init}$.
2: **repeat**
3:     Subdivide each cube of $\mathcal{L}$ congruently (*i.e.*, $\sigma \leftarrow \sigma/2$).
4:     **for** each cube $D_i$ in $\mathcal{L}$ **do**
5:        Calculate the rotation $\mathbf{R}_i = c(D_i)$ for the cube center.
6:        Compute lower $L(\mathbf{R}_i)$ and upper $U(\mathbf{R}_i)$ bounds.
7:     **end for**
8:     $L^* = \max_i L(\mathbf{R}_i)$, $i^* = \arg \max_i U(\mathbf{R}_i)$,
9:     $U^* = U(\mathbf{R}_{i^*})$, $\mathbf{R}^* = \mathbf{R}_{i^*}$.
10:     Remove all the cubes from $\mathcal{L}$ such that $U(\mathbf{R}_i) < L^*$.
11: **until** $\exists i$, such that $L(R_i) = U^*$ (*i.e.*, when at least one cube exists, whose lower bound is equal to $U^*$) or it reaches a desired accuracy level.
**Output:** $\mathbf{R}^*$ (*i.e.*, the rotation matrix maximizing the number of inliers).

### 4.2 Bounding Part

For the rotation search problem, a useful and efficient bound computation is suggested by Bazin *et al.* [22]. We directly restate the result of Bazin *et al.* for referring to the connection with our problem in Eq. (2).

**Proposition 1** (Bazin *et al.* [22]). *Given a cube $D$ with the half side length $\sigma$ and a mapping $c(D) = \bar{\mathbf{R}}$ that maps the cube $D$ to the rotation $\bar{\mathbf{R}}$ corresponding to the center of it, the following lower and upper bounds, $L_B$ and $U_B$, are valid for the inlier cardinality for any rotations in the cube $D$.*

$$L_B(\bar{\mathbf{R}}) = \max_{\{y_{ij}\}} \quad \sum_{i=1}^{N} \sum_{j=1}^{6} y_{ij}$$
$$s.t. \quad y_{ij} \angle(\mathbf{n}_i, \bar{\mathbf{R}} \mathbf{e}_j) \leq y_{ij} \tau,$$
$$y_{ij} \in \{0, 1\}, \forall i = 1 \cdots N, j = \{1 \cdots 6\}. \quad (3)$$

$$U_B(\bar{\mathbf{R}}) = \max_{\{y_{ij}\}} \quad \sum_{i=1}^{N} \sum_{j=1}^{6} y_{ij}$$
$$s.t. \quad y_{ij} \angle(\mathbf{n}_i, \bar{\mathbf{R}} \mathbf{e}_j) \leq y_{ij} (\tau + \sqrt{3}\sigma),$$
$$y_{ij} \in \{0, 1\}, \forall i = 1 \cdots N, j = \{1 \cdots 6\}. \quad (4)$$

*Proof.* Proof is deferred to [22]. □

In Proposition 1, the solutions of Eqs. (3) and (4), $L_B$ and $U_B$, are simply obtained by exhaustively checking the inlier constraint for each normal with respect to a given rotation $\bar{\mathbf{R}}$ that corresponds to the center of a given cube $D$. We refer to this method as `exhaustive BnB` to distinguish with our approach.

In the above bound computation, a single evaluation of lower and upper bounds has $O(N)$ complexity with respect to inlier computation. It is linear to the number of input normals, but along with the exponentially increased number of cubes by the branching step, this bound computation emerges as a main computational bottleneck in BnB frameworks.

### 4.3 Procedure of BnB for Rotation Search Space

The overall procedure of the BnB method for rotation search space is listed in Alg. 1. The algorithm reduces the volume of the search space iteratively by rejecting sub-spaces with the feasibility test until it converges to a globally optimal value or reaches the desired accuracy level.

To be specific, the cube-list $\mathcal{L}$ is first initialized with the cube $D_{init}$ that encloses the rotation ball $B_\pi$. At every iteration, each cube in the cube-list is subdivided into octal sub-cubes with the half length size of its parent cube and stored in the cube-list, while removing the parent cubes from the list. For each sub-cube, rotation center, upper and lower bounds are computed, and then feasibility test is conducted with the bounds.

In the feasibility test, with computed bounds and given a cube, the algorithm tests whether it is possible for the cube to contain a globally optimal solution. Specifically, when the cube is not in the feasible region of the solution space or when its upper bound is less than the current maximum lower bound $L^*$, we discard such a cube from the cube-list, as they are guaranteed not to contain the globally optimal solution. This makes infeasible sub-spaces excluded from search space at an early stage so that the global solution search can be done efficiently.

The remaining sub-spaces through the *bounding* step are subdivided recursively by *branching* step for further search. This procedure is carried out in every subdivision until a single cube remains, of which the lower bound and upper bound are same, or until the desired accuracy is achieved. The solution is the rotation of the center of the remaining cube that has the highest cardinality, *i.e.*, the rotation guaranteed to be global optimum.

## 5 EFFICIENT BOUND COMPUTATION

As aforementioned in Sec. 4.2, the computational bottleneck arises from counting the cardinality one by one. We avoid this exhaustive counting by transferring data to an efficient measurement space which is extended Gaussian image (EGI) [19]. On EGI, we relax the original problem defined in Eq. (2) to fully exploit a structural benefit of EGI, and propose new efficient bound functions. We describe the EGI representation at the most basic level in Sec. 5.1, then we introduce the new bound functions in Sec. 5.2.

### 5.1 Efficent Measurement Space (2D-EGI)

For EGI representation, we discretize a unit sphere (*i.e.*, *measurement space*) by simply tessellating it along azimuth and elevation axes, which have the ranges of $[0, 360°)$ and $[0, 180°]$, respectively. Then, a direction vector originated from the origin of the sphere is approximated and represented as its azimuth and elevation angles in a discrete unit. On the discretized azimuth and elevation coordinate, we construct a histogram of measurements (surface normals). As the histogram is known as an approximation to extended Gaussian image (EGI) representation [19], we will refer the EGI representation and the surface normal histogram interchangeably throughout this paper. The EGI representation on a 3D unit sphere can be directly transferred to the 2D-EGI domain (*a.k.a.* equirectangular projection), which provides powerful advantages as it allows the direct 2D representation [33] (especially with respect to memory access). The examples of EGI representation are shown as the heat map on the sphere (3D-EGI) and equirectangular map (2D-EGI) in the second column of Fig. 1. The notations of EGI are formalized as follows. We denote 2D-EGI space as $\mathtt{EGI} \in \mathbb{R}_+^{m \times n}$, where $m$ and $n$ are the height and width as $180 \times s$ and $360 \times s$, respectively, and $s$ denotes the EGI resolution parameter meaning that each axis unit is $1/s°$. There is a trade-off between accuracy and computation time by adjusting the EGI resolution.

When computing the bound values given a rotation (the center of a given cube), the inlier cardinality is counted for the normal vectors within the regions specified by a given inlier threshold (*e.g.*, $\tau$ or $\tau + \sqrt{3}\sigma$). We call this region an *inlier region*. In the case of 3D-EGI, following Proposition 1, an inlier region can be represented as a circular region as illustrated in Fig. 2a, where their boundaries of inlier regions are denoted by blue and red circles, which correspond to lower and upper bounds respectively. Let $\mathcal{X}_{\mathbf{L}} = \{\mathcal{X}_{L_j}\}_{j=1}^{6}$ and $\mathcal{X}_{\mathbf{U}} = \{\mathcal{X}_{U_j}\}_{j=1}^{6}$ be the sets of densely sampled boundary points of the lower and upper inlier regions around the six direction vectors $\{\mathbf{r}_j\}_{j=1}^6$ on the spherical manifold of the measurement space, where $\mathbf{r}_j = \mathbf{R}\mathbf{e}_j$ is a rotation axis.

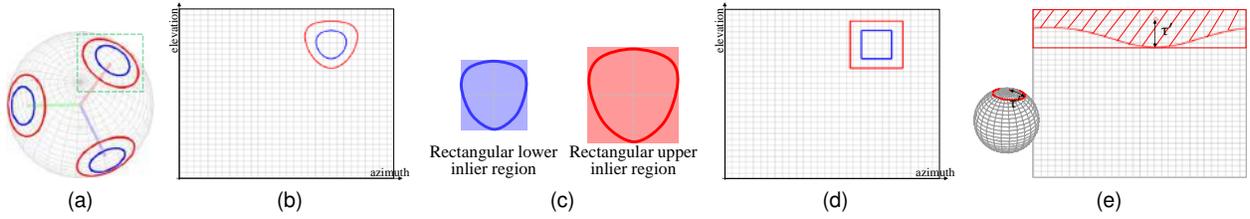

Fig. 2: Illustration of the geometric properties of the proposed efficient inlier regions: (a) Boundary sets $\mathcal{X}_\mathbf{L}$ (blue) and $\mathcal{X}_\mathbf{U}$ (red) of inlier region for lower and upper bounds of the original problem (*i.e.*, regions defined by Eqs. (3) and (4)) on a spherical domain. We visualize only three direction vectors out of six for illustration purpose. (b) An example of the transferred boundaries $\hat{\mathcal{X}}_{L_j}$ and $\hat{\mathcal{X}}_{U_j}$ on the 2D-EGI. (c) Relationship between the transferred boundaries and the rectangular boundaries. (d) An example of the boundaries of the rectangular bounds defined by Eqs. (6) and (7) on the 2D-EGI. (e) An example of the polar case (when a boundary includes a pole of the spherical domain), where $\tau'$ indicates an arbitrary threshold. In this case, the transferred boundary is distributed across all ranges of azimuth, and so it is for the rectangle boundary enclosing the transferred one's area.

Likewise, in 2D-EGI, the bound values can be computed by summing over the surface normal histogram within inlier regions, but in here the shapes of inlier regions are different from the ones on the 3D sphere. Once $\mathcal{X}_\mathbf{L}$ and $\mathcal{X}_\mathbf{U}$ are mapped onto the 2D-EGI, the transferred boundary sets $\hat{\mathcal{X}}_\mathbf{L}$ and $\hat{\mathcal{X}}_\mathbf{U}$ on the 2D-EGI domain appeared to have curved shapes as illustrated in Fig. 2b. The lower and upper bounds, $L_B$ and $U_B$, can be efficiently computed by summing the histogram values within the transferred boundary sets on the 2D-EGI, but still, it has a linear complexity with respect to the area of the curved inlier regions. We instead relax the problem, so that further speed-up is possible for the bound computation.

### 5.2 Rectangular Lower and Upper Bounds

Regardless of the tightness of the bounds, any valid bounds[5] guarantee global optimality in the BnB framework [17]. By slightly relaxing the inlier condition on the 2D-EGI domain, we can improve the computational efficiency significantly while preserving the globally optimal property if its bounds are still valid.

**Constraint relaxation**  Since the transferred boundary sets $\hat{\mathcal{X}}_\mathbf{L}$ and $\hat{\mathcal{X}}_\mathbf{U}$ on the 2D-EGI have non-linear shapes, exhaustively traversing within the transferred boundaries is mandatory for every computation of the bounds. We instead relax the problem defined in Eq. (2) by replacing the inlier constraint introducing non-linear shapes with new axis-aligned inlier constraints as:

$$\underset{\{y_{ij}\}, \mathbf{R} \in SO(3)}{\arg\max} \quad \sum_{i=1}^{N} \sum_{j=1}^{6} y_{ij}$$
$$\text{s.t.} \quad y_{ij}\phi(\mathbf{n}_i, \mathbf{R}\mathbf{e}_j) \le y_{ij}\tau_\text{el}, \quad (5)$$
$$y_{ij}\theta(\mathbf{n}_i, \mathbf{R}\mathbf{e}_j) \le y_{ij}\tau_\text{az},$$
$$y_{ij} \in \{0,1\}, \forall i=1 \cdots N, j=\{1 \cdots 6\},$$

where $\phi(\cdot,\cdot)$ and $\theta(\cdot,\cdot)$ are the angular distances between two vectors along the elevation and the azimuth axes of EGI, respectively, and $\tau_\text{el}$ and $\tau_\text{az}$ are inlier thresholds for each axis. We will discuss how to choose these inlier thresholds later. These constraints form a box constraint that can be efficiently computed.

---

[5]. A valid bound must satisfy the following conditions [17], [34]: 1) its lower and upper bounds satisfy inequality, *i.e.*, no solution in the search space yields smaller/larger inlier cardinality than that of lower/upper bound, 2) bounds asymptotically converge to the same value, *i.e.*, a gap between lower and upper bound eventually converges to zero as the branching steps are iterated. The formal mathematical validity of bounds will be discussed later.

Once we apply the relaxation, the problem we want to solve in Eq. (5) departs from the original one in Eq. (2). Thus, the globally optimal solution of the relaxed problem, the solution of Eq. (5), is not necessarily same with the one of `exhaustive BnB`, the solution of Eq. (2), but we expect it to be similar. This will be shown experimentally in Sec. 8.2.

**Rectangular bounds**  We apply BnB to the relaxed problem in Eq. (5) by defining new valid lower and upper bound functions similar to Proposition 1. For new bound functions to be closest to the original problem in Eq. (2), we find the tightest circumscribed rectangles of $\hat{\mathcal{X}}_\mathbf{L}$ and $\hat{\mathcal{X}}_\mathbf{U}$. We call these inlier regions the *rectangular inlier regions*, as shown in Fig. 2d. Once the boundaries are restricted to be axis-aligned and rectangle, the new boundaries are uniquely defined by finding the tightest circumscribed rectangle with four corner points along the elevation and azimuth axes, such that the rectangle includes the transferred boundary sets $\hat{\mathcal{X}}_\mathbf{L}$ and $\hat{\mathcal{X}}_\mathbf{U}$ as shown in Fig. 2c. Then, the bound functions of the relaxed problem in Eq. (5) can be defined as follows:

$$L_R(\bar{\mathbf{R}}) = \max_{\{y_{ij}\}} \quad \sum_{i=1}^{N} \sum_{j=1}^{6} y_{ij}$$
$$\text{s.t.} \quad y_{ij}\phi(\mathbf{n}_i, \bar{\mathbf{R}}\mathbf{e}_j) \le y_{ij}\tau_\text{el}, \quad (6)$$
$$y_{ij}\theta(\mathbf{n}_i, \bar{\mathbf{R}}\mathbf{e}_j) \le y_{ij}\tau_\text{az}^L(\phi(\mathbf{e}_2, \bar{\mathbf{R}}\mathbf{e}_j)),$$
$$y_{ij} \in \{0,1\}, \forall i=\{1 \cdots N\}, j=\{1 \cdots 6\},$$

$$U_R(\bar{\mathbf{R}}) = \max_{\{y_{ij}\}} \quad \sum_{i=1}^{N} \sum_{j=1}^{6} y_{ij}$$
$$\text{s.t.} \quad y_{ij}\phi(\mathbf{n}_i, \bar{\mathbf{R}}\mathbf{e}_j) \le y_{ij}(\tau_{el} + \sqrt{3}\sigma), \quad (7)$$
$$y_{ij}\theta(\mathbf{n}_i, \bar{\mathbf{R}}\mathbf{e}_j) \le y_{ij}\tau_\text{az}^U(\phi(\mathbf{e}_2, \bar{\mathbf{R}}\mathbf{e}_j)),$$
$$y_{ij} \in \{0,1\}, \forall i=\{1 \cdots N\}, j=\{1 \cdots 6\},$$

where $\bar{\mathbf{R}} = c(D)$ denotes the rotation corresponding to the center of the cube $D$, $\mathbf{e}_2$ is the $y$-axis directional basis of the 3D measurement space to measure an elevation angle. $\tau_\text{az}^L(\cdot)$ and $\tau_\text{az}^U(\cdot)$ indicate lower and upper inlier threshold functions for the azimuth axis, *i.e.*, each function returns the half width of the tightest circumscribed rectangle (circumscribing transferred boundary $\mathcal{X}_j$) at an elevation angle. Mathematically, transferred boundary $\mathcal{X}_j$, for a given direction vector and a threshold, on the 2D-EGI can be directly computed using Tissots indicatrix [35].[6] Using Tissots indicatrix, $\tau_\text{az}^L(\cdot)$ and $\tau_\text{az}^U(\cdot)$ can be computed in advance by finding each circumscribed rectangle of $\hat{\mathcal{X}}_\mathbf{L}$ and $\hat{\mathcal{X}}_\mathbf{U}$ and be stored into a look-up table as will be specified in Sec. 6.2.

---

[6]. Equirectangular projection is characterized with the parameter $(h=1, \theta'=90°)$ of Tissot's indicatrix model. Details refer to [35]. Specifically, its scale distortion rates along azimuth and elevation axes, $a$ and $b$, are $a(\phi)=\frac{1}{\cos\phi}$ and $b=1$, respectively, *i.e.*, the elevation axis is not distorted by the mapping, while the distortion along the azimuth axis is a function of elevation angle $\phi$.

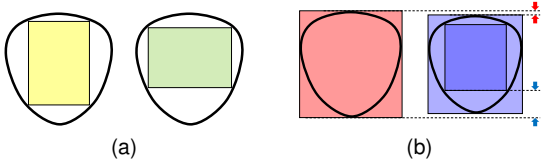

Fig. 3: Illustration of alternative ways to define a lower bound: (a) Examples of possible rectangular bounds inscribing the curve $\hat{\mathcal{X}}_{L_j}$, which are not uniquely determined. (b) Illustration of the gaps between upper (pink) and lower (blue) bound according to different lower bounds. The curved boundaries on the left and right represent the respective upper and lower curved boundaries, $\hat{\mathcal{X}}_{U_j}$ and $\hat{\mathcal{X}}_{L_j}$, at a level of subdivision. A gap between the two red arrows indicates a gap between the upper and the proposed lower bounds. The other one between the two blue arrows indicates a gap between the upper and lower bound inscribing the curve $\hat{\mathcal{X}}_{L_j}$, which cannot be converged.

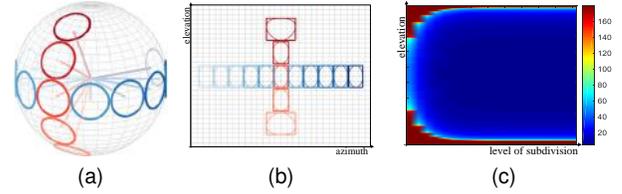

Fig. 4: Illustration of the properties of rectangular inlier region in the 2D-EGI domain for LUT: (a) The set of direction vectors and their inlier regions for the same threshold on sphere domain. The red-colored vectors have the same azimuth but different elevation, while the blue-colored vectors have the same elevation but different azimuth. (b) The rectangular inlier regions on the 2D-EGI. A width of the rectangular inlier region only depends on an elevation angle, while the height is fixed regardless of a location. (c) The visualization of matrix-form LUT for $\tau_{\mathrm{az}}^U(\cdot)$. We regard LUTs for each level of subdivision $k$ as column vectors, and concatenate them along the x-axis to form a matrix as shown in the figure. In the matrix-form LUT, the vertical axis indicates the elevation angle, the horizontal axis indicates the level of subdivision $k$, and a value of each entry encodes the half width size of its corresponding rectangular inlier region, which is color-coded.

As a polar case, when an inlier region (inner region of boundary points $\mathcal{X}_j$) includes pole, its transferred boundary $\hat{\mathcal{X}}_j$ maps across the whole azimuth range, as shown in Fig. 2e. Even in this case, the tightest circumscribed rectangle for the inner region of $\hat{\mathcal{X}}_j$ (slashed region in Fig. 2e) is uniquely determined and consistently provides validity of the rectangular bounds in Eqs. (6) or (7).

One may define an alternative lower bound of which form is a rectangular bound inscribing the curve $\hat{\mathcal{X}}_{L_j}$, instead of the circumscribed one. However, this definition has at least two flaws: 1) there are several ways to define inscribing rectangles, i.e., not uniquely determined as shown in Fig. 3a, and 2) as shown in Fig. 3b, the gap between the lower and upper bound does not converge to zero at any sub-division level, which is an essential condition for the validity of bounds [34], [36], [37]. Hence, the lower bound characterized by the inscribing rectangle may not guarantee global optimality as well as the convergence of the algorithm.

The feasibility constraints in the proposed rectangular inlier regions are interpreted as whether a pixel in the 2D-EGI map is inside the rectangle regions defined by inlier thresholds. Thus, the bound computation can be done by summing up the values in the rectangle regions, which is efficiently computed using the integral image [38], i.e. only with "+" and "−" operations of four corner values of the rectangular inlier region on the integral image.

### 5.3 Algorithm Procedure

With the proposed bound functions, the algorithm procedure is essentially same with Alg. 1 using breadth-first-search (BFS), except the bound computation in step 6 and stop condition of the algorithm. For the bounding step in the proposed scheme, the feasibility test is conducted based on the rectangular bounds (c.f. Sec. 5.2) instead of the bounds defined in Proposition 1. For the stop condition, we will derive a sufficient condition for algorithm termination later in Sec. 7.

## 6 IMPLEMENTATION DETAILS
### 6.1 Search Space Rejection

As aforementioned, given a rotation matrix $\mathbf{R} = [\mathbf{r}_1\ \mathbf{r}_2\ \mathbf{r}_3]$ of a cube, a direction $\mathbf{r}_i$ and its flipping direction vector $-\mathbf{r}_i$ (for $i = \{1, 2, 3\}$) indicate the same direction vector, i.e., the rotation matrices $\mathbf{R}$ and $-\mathbf{R}$ indicate completely the same solution in MF estimation. Hence, since we do not need to search within the other half-sphere regions in the rotation solution space, we exclude that half of the space.

### 6.2 Threshold Look-Up Table (LUT)

While the relaxed problem in Eq. (5) on the 2D-EGI provides computational efficiency, because of 3D to 2D-EGI mapping, rectangular inlier region on the 2D-EGI varies depending on the location of $\mathbf{r}_i$. In this mapping, we observed a few properties regarding the shape of the rectangular inlier region. Based on this observation, we relieve redundant and repetitive computation by pre-calculating thresholds as a look-up table (LUT). Note that we will represent azimuth and elevation values of $\mathbf{r}_i$ as $[r_i^\theta, r_i^\phi]$.

**Properties of inlier rectangular region** 1) Given $\mathbf{r}_i$ and $\tau_{el}$, the top elevation value of the rectangular inlier region is $\min(180°, r_i^\phi + \tau_{el})$, and the bottom one is $\max(0°, r_i^\phi - \tau_{el})$. We can simply calculate the height of a rectangular boundary by the difference of top and bottom elevation values, i.e., $\min(180°, r_i^\phi + \tau_{el}) - \max(0°, r_i^\phi - \tau_{el})$. 2) Except the polar case as shown in Fig. 2e, the half height of the rectangular inlier region is consistent as a constant, i.e., $\tau$ and $\tau + \sqrt{3}\sigma$ for the lower and upper rectangular inlier regions, respectively. That is, $\tau_{el}$ in the 2D-EGI is equal to $\tau$ in the original domain. Regardless of $\mathbf{r}_i$ locations, the heights of the rectangular inlier region are the same, as shown in Fig. 4b. 3) The width of rectangular inlier region depends only on the elevation angle $r_i^\phi$ of the mapping point of $\mathbf{r}_i$ (or $-\mathbf{r}_i$) on the 2D-EGI. Red-colored boxes in Fig. 4b show this property. 4) The shapes of the rectangular inlier regions on the same elevation angle are equal, regardless of the azimuth angles, as illustrated in Fig. 4b as blue-colored boxes.

Note that by virtue of the above properties 1) and 2), the value of $\tau_{el}$ can be directly used without any further conversion; thus, we do not have to calculate $\tau_{el}$ in advance. For the property 3), the relationship between the width and the elevation angle are related to a nonlinear distortion factor characterized by the notion of Tissot's indicatrix [35]. Thus, we pre-compute this value of $\tau_{az}(\cdot)$ according to all the elevation angles and sub-division levels.

For pre-computing $\tau_{az}(\cdot)$, we first generate a vector-shaped LUT, whose dimension is equal to the height (elevation unit) of the 2D-EGI, *i.e.*, $180 \times s$, for a threshold $\tau$. Each element of this LUT stores $\tau_{az}(\cdot)$, *i.e.*, the half width of rectangular inlier region of corresponding elevation angle. To calculate this value, we generate the set of boundary points of an inlier region, $\mathcal{X}_j$, for an elevation angle and a $\tau$, and map the set of points to the 2D-EGI to get corresponding transferred boundary $\tilde{\mathcal{X}}_j$. Then, we find the tightest circumscribed rectangle and store its half width value. We repeat this process for every elevation angle.

For $\tau_{az}^L(\cdot)$, we calculate the vector LUT for the original bound threshold $\tau$, because $\tau_{az}^L(\cdot)$ is invariant to the sub-division level. As for $\tau_{az}^U(\cdot)$, we generate the vector LUTs for a series of bound thresholds $\tau + \sqrt{3}\sigma_k$, where $\sigma_k = \sigma_0/2^k$ and $\sigma_0$ is the half side length of the initial cube. Note that by virtue of the subdivision rule of rotation search space [31], we can precompute a series of bound thresholds $\tau + \sqrt{3}\sigma_k$. By concatenating each vector-form LUT, we can generate a matrix form LUT (see Fig. 4c). By using the LUTs, we can obtain a rectangular inlier region without any repetitive computation while running the algorithm.

## 7 ANALYSIS

In this section, we present analyses of the proposed method in terms of computational complexity, properties of the proposed bounds, and its convergence behavior.

### 7.1 Computational Complexity

Proposition 1 [22] states that each bound computation has $O(N)$ complexity. For simplicity, let $C$ be the number of cubes that should be evaluated in the BnB framework until convergence. Then, the computation complexity of the whole procedure is $O(CN)$. This is the case of Hartley and Kahl [31] and Bazin *et al.* [22].

In our framework, constructing 2D-EGI by accumulating surface normals and the integral image take $O(N)$ and $O(H)$, respectively, only once at the initial stage, where $H$ denotes the number of bins of EGI, *i.e.*, $mn$. Also, since each bound computation with the 2D-EGI representation takes $O(1)$ on the integral image, the BnB procedure on the proposed problem shows a linear complexity with respect to the number of evaluated cubes $C$. Thus, the overall algorithm complexity is $O(C + H + N)$ which is still linear.

We can see that the proposed method is much faster than the `exhaustive BnB` method [22], which has a quadratic complexity $O(CN)$. In practice, given a single depth image with resolution $640 \times 480$, $N$ is around $300,000$ samples, and $C$ is the range of hundreds of thousands to millions (or even larger). This provides a sense of what makes the proposed algorithm near real-time.

### 7.2 Properties of Rectangular Bounds

Prior to analyzing the proposed BnB algorithm, we first show the validity of the proposed rectangular bounds. We have the following properties for the rectangular bounds.

**Lemma 1.** *The rectangular bounds defined in Eqs. (6) and (7) have the following properties:*

*(a) For a cube $D$, let $c_D^*$ be the optimal inlier cardinality of the relaxed problem (Eq. (5)) for $\mathbf{R} \in D$. Then the bounds $L_R$ and $U_R$ obtained by the proposed method satisfy $L_R \le c_D^* \le U_R$.*

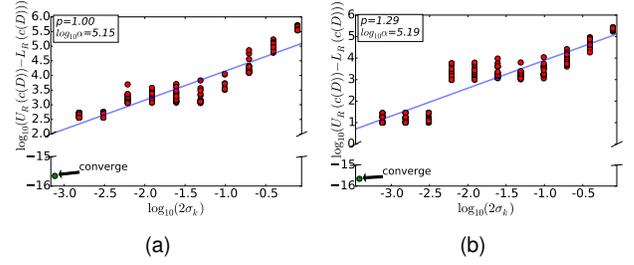

Fig. 5: Empirical convergence rates of the proposed bound function on (a) a synthetic example and (b) a real example of the NYUv2 dataset [10]. Although both the linear functions fit roughly well, depending on the levels of subdivision, both linear fit look under-fitted. It indicates the varying convergence rate depending on data.

*(b) As the maximum half side length $\sigma$ of a cube $D$ goes to zero by sub-divisions during BnB procedure in Alg. 1, there exists a small $\epsilon \ge 0$ such that $U_R - L_R \le \epsilon$.*

Because of space constraints, the proof can be found in the supplementary material. Lemma 1a shows that the maximum cardinality of a given cube always lies within $[L_R, U_R]$. Using this fact, we can test whether it is possible for a cube to have globally optimal cardinality in such a way comparing the upper bound of a cube with the maximum lower bound over all the cubes.

On the other hand, lemma 1b shows the convergence of the gap between the bounds. Thus, in an ideal case, we can guarantee to get point estimate as the number of iteration increases to infinity. These two lemmas imply that the proposed bound functions of Eqs. (6) and (7) are valid bounds in the perspective of BnB framework. These are indeed useful to further theoretically analyze the behavior of the rectangular bound in BnB, which will be discussed in the following section.

### 7.3 Convergence

From lemmas 1a and 1b, we can show that our algorithm converges to a globally optimal solution.

**Proposition 2.** *The BnB algorithm with the proposed rectangular bounds is guaranteed to converge to an $\epsilon$-optimal solution of the globally optimal solution. That is, once it converges to a point estimate ($\epsilon \to 0$), it guarantees to converge to a globally optimal solution.*

We only provide the sketch of proof here, and the detailed proof can be found in the supplementary material. In the case of $s \to \infty$ (continuous EGI), the classic convergence result of BnB with the proposed bounds, $L_R$ and $U_R$, is valid by lemmas 1a and 1b and by following the results of [17], [34]. That is, the proposed method guarantees a globally optimal solution once it converges to a point estimate. This is an ideal case in that $s \to \infty$. Now, consider $s$ is finite (discrete EGI). Let $\sqrt{3}(\sigma_k - \sigma_{k+1}) \le 1/s^\circ$ be a stop criterion, where $k$ is the level of subdivision. Since the stop criterion is a sufficient condition for the convergence in the discrete case, we can prove that geodesic angular distance between the globally optimal solution and a solution terminated with the stop criterion can be upper-bounded by $\epsilon$ expressed by $1/s^\circ$ in the right term of the stop criterion.

Now, we might analyze the convergence rate of the proposed algorithm, which is rather useful to understand an algorithm. The convergence rate of a BnB algorithm [39] can be defined as **Definition 1** (Convergence rate). *Let $D \subset SO(3)$ be a cube and a function $f:D \to \mathbb{R}$. Furthermore, consider the maximization problem, $\max_{\mathbf{R} \in D} f(\mathbf{R})$. We say a bound function has the convergence rate $p \in \mathbb{N}$ if there exists a fixed constant $\alpha > 0$ such that*

$$U\left(c(D')\right) - f\left(D'\right) \leq \alpha \cdot diameter(D')^p \quad (8)$$

*for all cubes $D' \subset D$.*

Unfortunately, in BnB, most of the research on convergence rate of bound functions has been done only for analytic functions [39], [40], [41], while our bound functions in Eqs. (6) and (7) are a purely non-parametric and non-analytic form that highly depends on data.[7] It is quite challenging to formalize in that, according to data, the convergence rate will be varying.

Instead, as presented in Scholz [40] and Schöbel and Scholz [39], we measure empirical convergence rates to understand the behavior of the proposed algorithm. Following the method of Schöbel and Scholz, to obtain empirical $\log \alpha$ and $p$, we fit a linear regression of the following equation:

$$\log \left\{U_R\left(c(D)\right) - L_R\left(c(D)\right)\right\} = \log \alpha + p \log 2\sigma_k \quad (9)$$

for randomly sampled cubes during the BnB procedure. Fig. 5 plots two empirical results *w.r.t.* synthetic and real examples. According to data statistics, overall convergence behaviors of Figs. 5a and 5b including slopes are different. Moreover, depending on the levels of subdivision, the linear fits seem to be under-fitted, *i.e.*, the convergence rates vary according to the level. These phenomena may come from two sources: 1) data characteristics of MF estimation, and 2) non-analytic property of the proposed bound function. Nonetheless, Fig. 5 shows that our algorithm has the empirical rate of convergence $p \approx 1$, which converges approximately linearly.

## 8 EXPERIMENTAL RESULT

In this section, we present our experimental results to explore the performance of the proposed method and to compare with other recent approaches. We systematically experiment with several criteria in simulation in Sec. 8.1, which shows how sensitive, accurate and robust the proposed algorithm is over other competitive methods. Additionally, we analyze the convergence property and time profile of the proposed method. Then, in Sec. 8.2, we further evaluate our method on real datasets, the NYUv2 dataset [10] and the sequential dataset [4], which shows the practicality of the proposed method in terms of accuracy and speed. Lastly, we present various extensions of our core algorithm to several applications such as multiple Manhattan frame estimation, 3D rotation based video stabilization, and vanishing point estimation (line clustering) in Sec. 8.3. A large-scale experiment is involved in the multiple Manhattan frame estimation application of Sec. 8.3.

### 8.1 Simulation

In the synthetic simulation, we perform various experiments to demonstrate stability (accuracy, robustness, and convergence) and efficiency (time profile) of the proposed BnB method. For synthetic data, we randomly drew three direction vectors orthogonal to each other, which correspond to ground truth MF, and additional direction vectors as outlier directions (we choose at least three

7. It is also the same for exhaustive BnB.

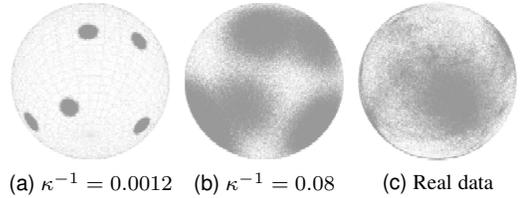

(a) $\kappa^{-1} = 0.0012$  (b) $\kappa^{-1} = 0.08$  (c) Real data

Fig. 6: Distributions of surface normals: (a) and (b) are synthetic data distribution according to $\kappa^{-1}$ of vMF, $0.0012$ and $0.08$, respectively. (c) A sample distribution of real data from the NYUv2 dataset [10].

outlier directions but not orthogonal to each other). We then sampled $300,000$ surface normals on von-Mises-Fisher (vMF) distribution [42], [43],[8] which is an isotropic distribution on a unit sphere. We also uniformly sampled $20,000$ surface normals on the unit sphere to generate sparse noise (see Fig. 6). Unless specifically mentioned, we fix the inlier threshold $\tau$ as $5°$, the EGI resolution parameter $s$ as 2, and the inverse kappa $\kappa^{-1}$ as $0.01$ for each experiment.

As a metric, given a ground truth MF (rotation matrix $\mathbf{R}_{gt}$) and an estimated MF (rotation matrix $\mathbf{R}_{est}$), the angular deviation $\boldsymbol{\theta}$ between the two matrices is measured by

$$\boldsymbol{\theta} = arccos(max(abs(\mathbf{R}_{gt}^\top \mathbf{R}_{est}))), \quad (10)$$

where the entries of $\boldsymbol{\theta} = [\theta_x, \theta_y, \theta_z]^\top$ are angular error deviations of each axis, $abs(\cdot)$ is element-wise absolute operation and deals with flipping directions, $max(\cdot)$ is row-wise max operation, *i.e.*, it returns max value of each row, and $arccos(\cdot)$ is element-wise arc cosine operation. Eq. (10) compares each axis of $\mathbf{R}_{gt}$ with three axes of $\mathbf{R}_{est}$ and computes smaller deviation of each axis. We ran each experiment 50 times on MATLAB and measured the mean of the max angular error.

**Accuracy** We evaluate how much accuracy of our method is affected by factors such as the EGI resolution parameter, the inlier threshold, and data distribution. According to the EGI resolution parameter $s$, we test the trade-off between accuracy and runtime of the proposed method. Figure 7a shows that, as $s$ increases, the accuracy is improved, while the increasing rate of the runtime is critical. In other experiments, if not mentioned, we empirically choose $s = 2$ as a reasonable trade-off. As another line of experiment, Fig. 7b shows that our method has stable accuracy regardless of the inlier threshold $\tau$.

We also evaluate accuracy according to the variance parameter of vMF distribution, namely $\kappa^{-1}$, and compare our method against recent MF estimation methods: MMF [2], RMFE [3], and RTMF [4].[9] In Straub *et al.* [4], three types of RTMF were proposed: vMF-based, approximate, and tangent-space Gaussian model. Since vMF-based RTMF (vMF-RTMF) provides the best quantitative performance among them, we used it as the representative RTMF and refer to as RTMF simply, unless specified. For this experiment, we test $\kappa^{-1}$ on the range of $[0.0012, 0.08]$ in a log-scale. In Fig. 7d, each color dot indicates its mean of the maximum angular error, and each bar represents the standard deviation of

8. The von Mises-Fisher distribution for the $p$-dimensional unit vector $\mathbf{x}$ is defined by $f_p(\mathbf{x};\mu,\kappa) = Z_p(\kappa)\exp(\kappa\mu^T\mathbf{x})$, where $\mu$ is a mean direction, $\kappa$ is a concentration, and normalization constant $Z_p(\kappa) = \kappa^{p/2-1}(2\pi)^{-p/2}\mathbf{I}_{p/2-1}(\kappa)^{-1}$, where $\mathbf{I}_v$ denotes the modified Bessel function of the first kind at order $v$.

9. For a fair comparison, we used the publicly available codes and original parameters of the original authors.

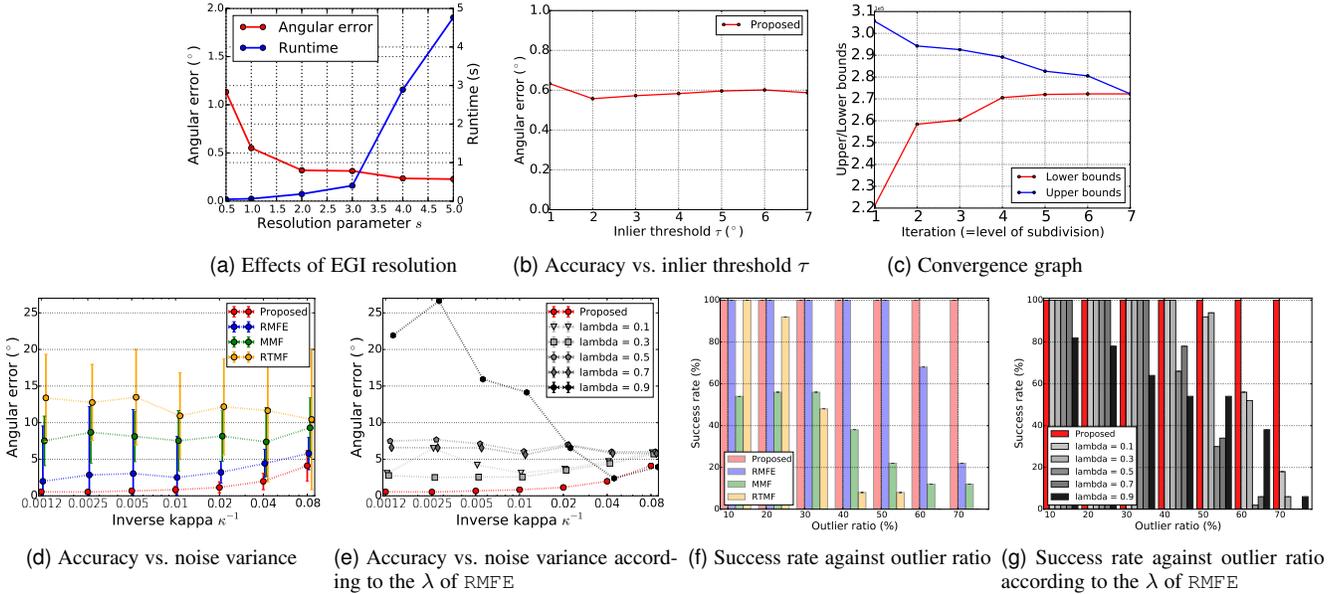

Fig. 7: Simulation results for observing the behaviors of the proposed method.

maximum angular error. MMF and RTMF estimation show relatively large errors with unstable standard deviations. While RMFE shows less error than those of MMF and RTMF, it also has unstable deviations due to the bias toward an outlier direction. Even with increasing $\lambda$ value,[10] RMFE shows unstable accuracy, as shown in Fig. 7e. On the other hand, our method shows stable and accurate performance.

**Robustness** We validate the robustness of our method against outliers by comparing success rates with other methods. For outlier generation, we vary a ratio of outlier surface normals from $10\%$ to $70\%$ for a fixed kappa value (we set $\kappa=128$), and assign outlier normals to one of the aforementioned outlier directions to be clustered. We classify a success of each trial if angular error against ground truth is lower than $5°$. As shown in Fig. 7f, MMF and RTMF show low success rates and rapid decrease as the outlier ratio increases. RMFE, on the other hand, shows relatively stable success rate, but it is also sensitive in the high outlier ratio regimes regardless of $\lambda$, as shown in Fig. 7g. On the other hand, the proposed method turns out to be robust even in the high outlier ratio regimes.

**Convergence** Fig. 7c shows how many sub-divisions (iterations) are typically required until convergence.[11] It commonly converges within 7 iterations. Moreover, it empirically shows that the lower and the upper bounds of the proposed BnB method converge to a specific value (implying the validity of the bounds), which supports the convergence property in Proposition 2.

**Time profiling** To show the computational efficiency of our method, we compared the time profiles of exhaustive BnB [22] and the proposed BnB in Fig. 8. Both methods have three steps in common: branch (subdivision), rotation center estimation, and bound computation, while the proposed BnB has an additional EGI generation step for efficient bound estimation. For exhaustive BnB, bound computation takes $108.178s$, which is $99.98\%$ of the entire computational time, while the counterpart of the proposed BnB takes only $4.6ms$. This turns out to reduce the bound computation time by more than 20,000 times.

---

10. In RMFE [3], $\lambda$ is a weight parameter for sparsity.

11. With the breadth-first-search (BFS), the terminologies, *iterations* and *subdivisions* are used in the same meaning.

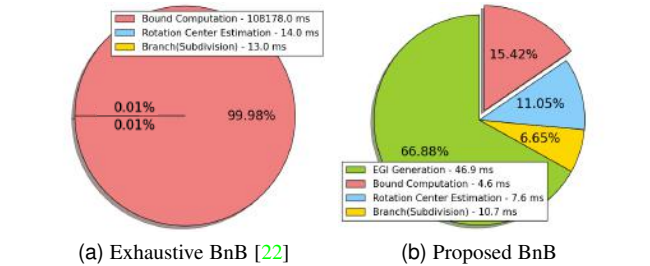

(a) Exhaustive BnB [22]     (b) Proposed BnB

Fig. 8: Time profiles of the exhaustive BnB [22] and the proposed BnB approaches.

| Method | MPE | MMF | ES | RMFE | RTMF | Exhaustive BnB | Proposed |
|---|---|---|---|---|---|---|---|
| $\theta_x$ | 26.3° | 5.3° | 2.3° | 2.3° | 4.1° | 2.9° | 3.0° |
| $\theta_y$ | 18.1° | 4.6° | 5.6° | 4.7° | 2.7° | 1.8° | 2.0° |
| $\theta_z$ | 18.2° | 5.3° | 2.9° | 2.8° | 3.9° | 2.8° | 2.9° |
| Avg. | 20.87° | 5.07° | 3.60° | 3.27° | 3.57° | **2.5°** | 2.63° |
| Runtime (s) | 2.8 | 148.7 | 21.4 | 0.9 | - | 117.06 | **0.07** |

TABLE 2: Comparisons of average angular error and runtime on the NYUv2 dataset [10] with the evaluation protocol of Ghanem *et al.* [3].

### 8.2 Real-World Experiment

**Evaluation on the NYUv2 dataset [10]** We evaluate the proposed method in terms of accuracy and speed with the real-world benchmark, the NYUv2 dataset, which contains $1,449$ RGB-D images of various indoor scenes. In particular, we used the recently introduced MF evaluation protocol [3] on the NYUv2 dataset for quantitative evaluation. From each scene of the NYUv2 dataset, we extract around $300,000$ surface normals as an input. We fix all the parameters of the proposed method as before. In Table 2, we compare exhaustive BnB and the proposed BnB with the recent MF estimation approaches: MPE [28], MMF [2], ES [10], RMFE [3], and RTMF [4]. In case of MMF, we run 10 iterations, calculate the angular errors for multiple cases, and choose a minimum error to alleviate the randomness of its initialization. For RTMF, we do not compute runtime since it utilizes GPU parallelization. We test average angular errors on similar hardware configurations (*i.e.*, a 3GHz workstation on MATLAB) as done by Ghanem *et al.* [3]. For MPE, ES, and RMFE, we directly quote the results from Ghanem *et al.*

Since MPE is based on the assumption that a significant portion of the scene consists of the floor plane, it is sensitive to scene clutter and outliers. `MMF` shows less accurate result than those of `ES` and `RMFE`. We deduce the reason for `MMF`'s poor performance in angular errors as the absence of noise/outlier handling. `ES` and `RMFE` show comparable results, but their runtime is inefficient for real-time applications. While `RTMF` shows a reasonable result with real-time efficiency, it heavily relies on GPU requirement. `Exhaustive BnB` shows the most accurate result, but its runtime is intractable in the efficiency aspect. On the other hand, the proposed BnB performs stable while achieving near real-time efficiency. The accuracy difference between the `exhaustive BnB` and the proposed one may come from the relaxation gap, in that we solve a relaxed problem defined in Eq. (5), which is discussed later.

**Evaluation on the sequence data [4]** As a quantitative evaluation for a sequence dataset, we used the video sequence dataset provided by Straub *et al*. [4] of which ground truth is captured by using a Vicon motion capture system to track a 3D pose of an RGB-D sensor with an attached IMU sensor. This dataset includes randomly waving camera motions in front of a wall with a rectangular pillar. Please refer the details of this dataset in Straub *et al*.

For the sequence scenario, Straub *et al*. [4] proposed a temporal information fusion approach by extended Kalman filter (EKF) with IMU sensor data, which is combined with `RTMF`'s variants: vMF-based and approximate `RTMF`s (*i.e.*, `vMF-RTMF` and `approx.-RTMF`). We compare our method without any modification against their methods.

For the dataset of 90 s length, Fig. 10 shows that the angular RMSEs of `vMF-RTMF` and `approx.-RTMF` are $6.39°$ and $4.92°$, respectively. Using the EKF fusion with the IMU sensor, the angular RMSEs of `vMF-RTMF` and `approx.-RTMF` are $3.05°$ and $3.28°$, respectively.[12] On the other hand, the angular RMSE of our method is $\mathbf{1.53°}$. Note that even though our method is applied frame-by-frame independently without any temporal consistency, our method outperforms all versions of `RTMF`.

### 8.3 Applications

We present several applications of our method: multiple Manhattan frame estimation, 3D rotation based video stabilization, and vanishing point estimation (line clustering). These are built on the robustness of the proposed method as evaluated in the previous sections, and demonstrate the practicality of our method.

**Extension to multiple Manhattan frames (MMFs)** Since the strict MW assumption cannot represent general real-world indoor and urban scenes, Straub *et al*. [2] introduced a more flexible description of a scene, consisting of multiple MFs, known as a mixture of Manhattan frames. As an application, we extended the proposed method to MMFs by sequentially finding different MFs.

To estimate MMFs, we apply a greedy-like algorithm. For a given input data, we estimate the best optimal MF by the proposed BnB and update the normal data by discarding the set of normals that correspond to the inliers of the optimal MF. We then sequentially estimate the next optimal MF given the updated normal data.[13] Following Straub *et al*. [2], we only regard MFs of which inlier normals are more than $15\%$ of all valid

| Method | Straub *et al*. [2] | | Proposed | |
|---|---|---|---|---|
| MFs case | 1 MF | 2 MFs | 1 MF | 2 MFs |
| $\theta_x$ | 5.2° | 13.8° | 2.8° | 6.3° |
| $\theta_y$ | 4.6° | 16.2° | 1.9° | 10.3° |
| $\theta_z$ | 5.2° | 16.5° | 2.7° | 10.4° |

TABLE 3: Quantitative evaluation of multiple MFs on the NYUv2 dataset [10].

normals as good MF estimates, to alleviate poor quality of depth measurements.

We demonstrate our extension, *i.e.*, MMFs inference for the small (indoor) and large (outdoor) scale datasets in Fig. 9. For the small-scale dataset, we again used the NYUv2 dataset [10] for qualitative and quantitative evaluation. The proposed MMF inference shows qualitatively reasonable results, as shown in Figs. 9a and 9b. For the quantitative evaluation of MMFs, we newly provide an additional evaluation protocol on the NYUv2 dataset, *i.e.*, ground truth of MMFs, where we annotate MMF ground truth by combining the ground truth number of multiple MFs from Straub *et al*. [2][14] and 3D bounding boxes of scene objects from the SUN RGB-D dataset [44].[15] We describe the detailed procedures for the evaluation protocol in the supplementary material. Among $1,449$ scenes of the NYUv2 dataset, we used 173 scenes that are categorized as multiple MFs. Using the metric in Eq. (10), we compute every possible combination between an estimated MF and all ground truth MFs and choose the smallest deviation as a rotation error of an estimated MF and compare with `MMF` [2] in Table 3. In both 1 MF and 2 MFs cases, our method outperforms `MMF`. Interestingly, in case of 2 MFs, both methods show relatively larger angular error than those of 1 MF. This is because both methods commonly estimate multiple MFs that do not share any axis, and ground truth MFs in the dataset usually share a single vertical axis (typically orthogonal to the ground plane) as a dominant axis. Thus, results of both methods for 2 MFs are relatively sensitive than those of 1 MF.

For the large-scale test, we use the point clouds captured in the city center of Bremen, Germany, from Robotic 3D scan repository,[16] which freely provides 3D point clouds captured by the laser scanner and pose data in various scenes. This dataset consists of around 81 million 3D points. For this dataset, our method takes $10.124$ s to estimate MMFs in total. In the breakdown in timing, the EGI generation takes $9.765$ s, which occupies most of the computational time. As shown in Fig. 9c, the proposed method properly results in 4 MFs in the scene.

**3D rotation based video stabilization** The goal of video stabilization is to generate a visually stable and pleasant video from a jitter video by camera shake. Recently, Jia *et al*. [46] proposed a video stabilization that exploits a 3D rotational motion model of a camera based on the fact that 3D motion models can reflect more realistic motion than 2D affine or projective motion.[17] For a 3D rotational model, they used a built-in gyroscope of smartphones or tablets. In our scenario, instead of the rotations obtained by a gyroscope, we utilize MF obtained by our method

---

12. We directly quote the result of `RTMF` [4].

13. Since we do not estimate all MFs jointly, our algorithm for the MMFs application does not guarantee global optimality for whole MFs.

14. http://people.csail.mit.edu/jstraub/_pages/nyu-mmf-dataset/

15. The SUN RGB-D dataset [44] consists of the union set of several datasets including the NYUv2 dataset (acquired by Kinect v1) as well as other datasets acquired by four different RGB-D sensors.

16. http://kos.informatik.uni-osnabrueck.de/3Dscans/

17. As the stabilization method used for this experiment only optimizes a given rotation path while ignoring a projective motion between image sequences, the stabilization performance solely depends on the quality of rotations estimated.

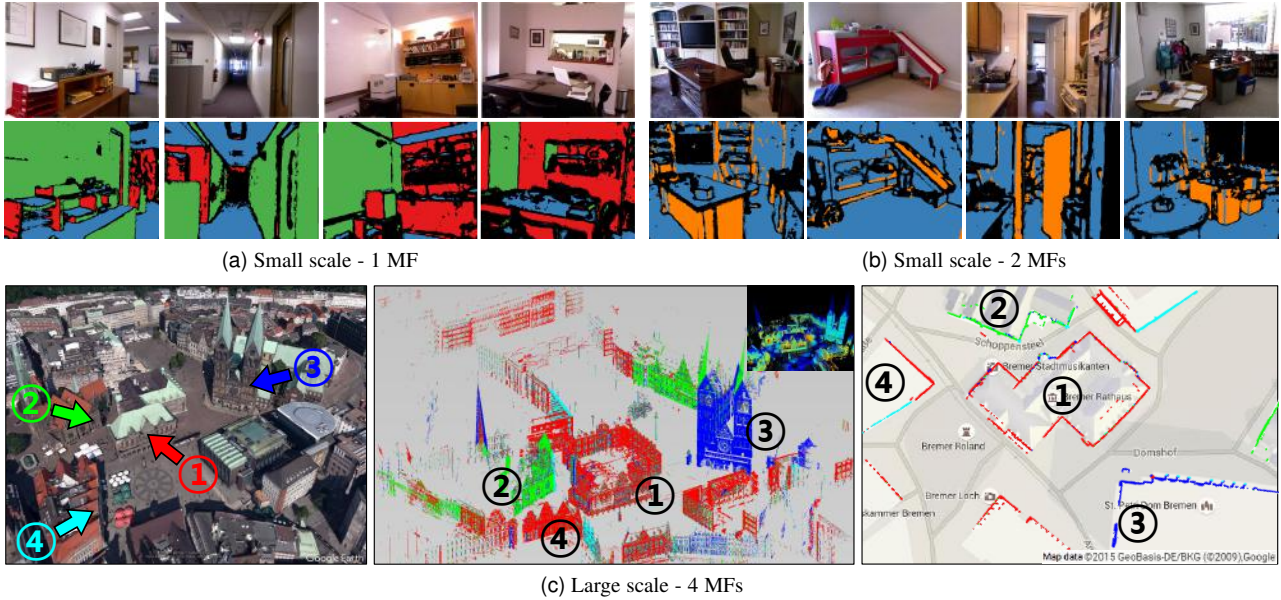

Fig. 9: Multiple Manhattan frames (MMFs) results for small- and large-scale datasets: (a) Examples of single MF (each RGB color indicates respective MF axes). (b) Examples of two MFs (orange and blue colors indicate respective MF). (c) Large scale example, where we estimated 4 MFs, and colored and numbered respective MFs to be clear. The NYUv2 indoor scenes dataset [10] is used for (a,b), and for (c), images of the city center of Bremen, Germany, are used by courtesy of the Google map and Google earth.

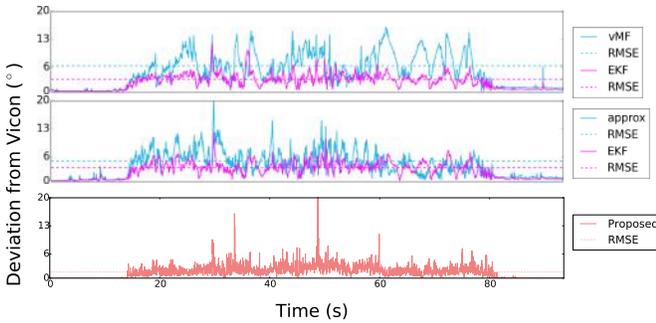

Fig. 10: Angular error comparison for the sequential ground truth. [Top, Middle] Respective `vMF-RTMF` and `approx.-RTMF` with/without EKF fusion with the IMU [4]. [Bottom] The proposed method. RMSE: `vMF-RTMF` (6.39°), `vMF-RTMF+IMU` (3.05°), `approx.-RTMF` (4.92°), `approx.-RTMF+IMU` (3.28°), and ours (1.53°).

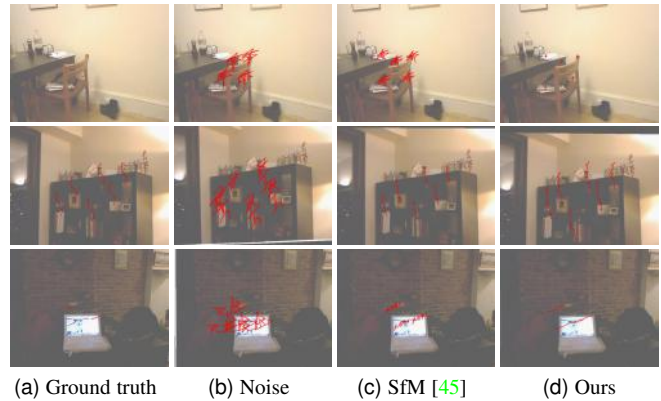

Fig. 11: 3D rotation based video stabilization. We overlay the feature trajectories for qualitative visualization purpose: (a) Ground truth motion. (b) Motion of the input sequence perturbed by synthetic noisy motion. (c) Motion of stabilized frames by rotations obtained by a conventional SfM [45]. (d) Motion of stabilized frames by rotations obtained by our method. Given rotations of frames, we use the stabilization algorithm proposed by Jia *et al*. [46]. We tested four scenes of which motions are different types: from the top to bottom row, living room 0002, living room 0006, and living room 0009 of the NYUv2 dataset [10].

as 3D rotation matrices, so that the applicability of our method on video stabilization is verified. Note that we only use depth information to estimate 3D rotation matrix. Namely, as inputs for the stabilization [46], we only use estimated MF to obtain smoothed camera path, and applied to RGB images.

We experiment on videos of the NYUv2 dataset [10]. We added rotation noise to mimic egocentric head motions and applied to the image and depth sequences. For visualization purpose, following the way of stabilization comparison in [46], we overlay sampled feature trajectories, which are tightly related to the stabilization performance (see Fig. 11). The initial features are detected by FAST [47], then be tracked by KLT [48] along consecutive frames. Note that these trajectories are not used in our algorithm, but only for visualization purpose.

Fig. 11a shows the smooth feature trajectories of the original video (ground truth), which represent the true camera motion. Fig. 11b shows the jittering feature trajectories by synthetic rotation noise applied to the original videos. We compare the feature trajectories of the videos processed by stabilization [46] using 3D rotations estimated by our MF estimation (Fig. 11d) and 3D rotations estimated by conventional structure from motion (SfM) [45] (Fig. 11c). Overall, stabilization based on our approach produces more similar motion with ground truth when compared with those based on SfM. SfM-based stabilization provides comparable results when objects and textured planes are dominant, but inherently SfM does not work for the scenes with a homogeneous plane. Also, as SfM runs on whole image sequences (matching every image pair with bundle adjustment optimizer), it cannot run online and takes significant time. One exceptional case to note is shown at the second row in Fig. 11. Both our approach and SfM show straight motion while the ground truth motion is wavy. It is because the stabilization optimizes the rotation path to be smooth, deforming the original wavy motion. Nonetheless, the stabilization using our method shows plausibly stabilized results.

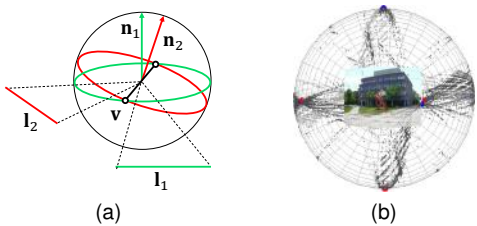

Fig. 12: Illustration of VP candidate $\mathbf{v}$ in 3D space (Gaussian sphere) and an example of VP candidates (gray points) for the York urban dataset (YUD) [51]. Red and blue dots indicate ground truth VP and estimated VP, respectively.

**Vanishing point estimation and line clustering** Vanishing point (VP) estimation and line clustering are well-known chicken-and-egg problems [30]; VPs can be obtained if the line clustering is known, and line clustering can be done if the VPs are known. Based on this fact, we first estimate VPs using our MF estimation and then line clustering as applications. In this scenario, we demonstrate the direct applicability of our method to VP estimation on structured scenes. In structured scenes, VPs in a 3D space (*a.k.a.*, vanishing directions, VDs) tend to be orthogonal to each other. Under the MW assumption, we simply formulate the estimation of orthogonal VPs as consensus set maximization in the same way with MF estimation: estimating dominant and orthogonal directions (*i.e.*, orthogonal VPs) given measurements (*i.e.*, VP candidates). For simplicity to show direct applicability, we assume the known camera intrinsic.

Prior to results, we first explain how we generate VP candidates in a 3D space. We introduce a concept of the Gaussian sphere [49], which is a convenient way to represent an image when the camera calibration parameters are known. On the Gaussian sphere, a line $\mathbf{l}_i$ in the image is back-projected onto the sphere as a great circle, which is represented by a unit normal vector $\mathbf{n}_i$ (see Fig. 12a). Unless two lines are exactly same, two great circles of two lines intersect at two points, which are antipodal points [50]. They correspond to the VP candidate and are computed by $\mathbf{v} = \mathbf{n}_i \times \mathbf{n}_j$, where $\mathbf{n}_i$ and $\mathbf{n}_j$ are the normal vectors of two lines. We calculate VP candidates for every combination of two lines and use those as input measurements of EGI for our VP estimation. Fig. 12b is an example for calculated 3D VP candidates.

For validation, we use the York urban dataset (YUD) [51] consisting of 102 images acquired in indoor and outdoor man-made environments. The results in Fig. 13b qualitatively show that our method successfully finds VPs and line clustering in the real-world dataset. For a quantitative evaluation for the YUD, following the evaluation protocol [52], [53], we evaluate our VP estimation using the cumulative density function of the angle to the horizon and its area under curve (AUC), as shown in Fig. 13a. Among various VP estimation methods [52], [53], [54], [55], [56], [57], we compare our method against the recent state-of-the-art methods that measure the horizon error. Also, we compare with RMFE [3] as a reference method in that the method estimates VPs in the same 3D domain as ours.

In 2D horizon error, our method does not perform better than the recent methods, which are specifically designed to have minimum horizon error of 2D VP estimation. Comparing with RMFE on the same domain, our method outperforms RMFE, which is a consistent result with other experiments. On comparing with the recent methods, the rationale behind the limited performance of our method may be that, different from the recent methods that directly work on the 2D domain, we generate VP candidates on the Gaussian sphere from a set of 2D lines and estimate optimal VPs in a 3D space. This unavoidable process may introduce another source of error. Another potential limit may come from the orthogonality constraint of our method, while the other 2D VP estimation methods have some flexibility. Although our method is not the state-of-the-art on VP estimation, the applicability of our method to VP is quantitatively validated through this benchmark. Regarding the global optimality, our method may be useful for the scenarios requiring stability, efficiency, and robustness.

## 9 DISCUSSION AND CONCLUSION

Based on the motivation that MF estimation requires two properties (*i.e.*, stability and efficiency) for general performance, we have presented a robust and fast MF estimation that guarantees a globally optimal solution. This can be achieved by relaxing the original problem and computing the bounds efficiently on the 2D-EGI domain. By virtue of this, the complexity of the bound computation is dramatically reduced to constant complexity. In theoretical aspects, we show the validity and the convergence of the proposed bounds. Then, based on these, the convergence property to global optimality is established. To validate the stability and the efficiency of our method, we systematically demonstrate our method on various synthetic and real-world datasets. These show that our method outperforms other state-of-the-art methods for speed, accuracy, and robustness. Since MF estimation is a fundamental problem, our proposed method, which is efficient, robust, and accurate, may improve the stability of other subsequent applications that adopt ours as preprocessing, including what we have shown: multiple MF estimation, video stabilization, and vanishing point estimation (line clustering). In the following, we discuss some open questions related to our paper:

**Relaxation gap** By relaxing the original problem in Eq. (2), our problem in Eq. (5) departs from the original one; hence, globally optimal solutions of both are not necessarily same, which introduces a relaxation gap between two problems. This gap leads to the accuracy difference between exhaustive BnB and our method in Table 2. Fortunately, the difference may be negligible, but still, further mathematical analysis of the relaxation gap would be a worthwhile and interesting research direction.

**Alternative tree search method** One can adopt other tree search techniques like depth-first-search (DFS) or A* [58], so that by identifying a promising direction for finding a higher value of lower bounds as early as possible, an efficient traverse can be done by inducing early rejection frequently. Since our contribution is on an efficient measurement computation, for simplicity and fairness, we use BFS for rotation solution search, which has been used as a standard technique for many BnB literatures [22], [30], [59] and its implementation and memory management are relatively easy. Hence, once BFS is replaced with an alternative technique, the overall efficiency will be improved. We leave this to future research.

## ACKNOWLEDGMENTS

The authors appreciate helpful discussion and comments from Prof. Jinwoo Shin and Gilwoo Lee. This work was supported by the Technology Innovation Program (No. 2017-10069072) funded By the Ministry of Trade, Industry & Energy (MOTIE, Korea).

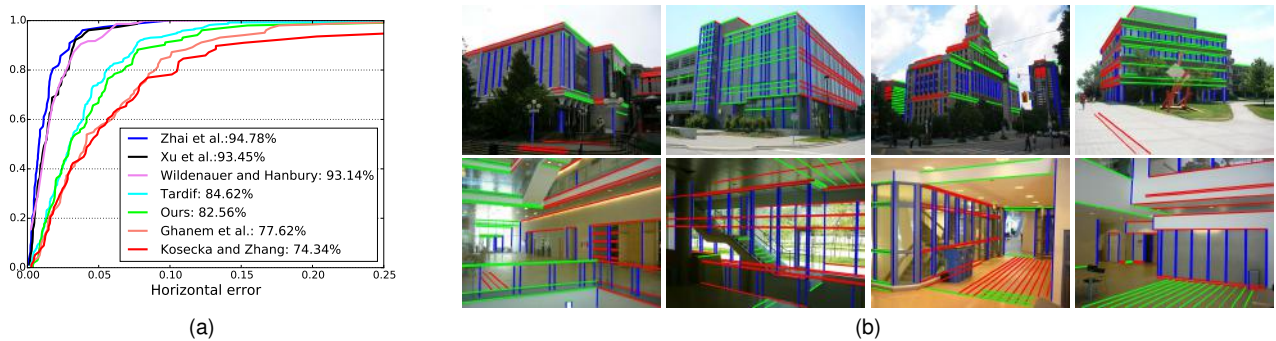

Fig. 13: The result of VP estimation and line clustering: (a) Cumulative histograms for the horizon error and its AUC for the York urban dataset [51]. The fraction of images (y-axis) with a horizon error less than a threshold (x-axis). The legends are sorted based on AUC. (b) Examples of line clustering obtained by our method on the York urban dataset. The same colored lines indicate a cluster associated with the same VP.

## OVERVIEW OF SUPPLEMENTARY MATERIAL

Here, we present additional details of the proposed evaluation protocol and a benchmark MMF ground truth (GT) dataset, and theoretical proofs that could not be included in the main text because of space constraints.

More specifically, we describe which kinds of information are utilized and how these are combined to properly build a new multiple MF ground truth for the NYUv2 dataset in Sec. 10. Also, we introduce several relationships necessary for smooth proof in Sec. 11.1, and prove Lemmas 1a and 1b that theoretically support the global optimality and convergence of the proposed method in Secs. 11.2, 11.3 and 11.4.

All the numberings for references, figures, tables, and equations are consistent with the main manuscript.

## 10 MMFs GROUND TRUTH FOR THE NYUv2

The NYUv2 dataset [2] is a commonly used RGB-D indoor benchmark dataset in computer vision fields: object segmentation [60], object detection [61], and depth estimation [62]. On top of the NYUv2 dataset, other researchers have extended benchmark protocols to other problems:

1) **Annotations for the number of multiple MFs**: Straub *et al.* [2] annotated the number of multiple MF GTs for each scene in the NYUv2 dataset for *qualitatively* demonstrating their inference performance for the number of MFs.[18]
2) **Single dominant MF GT**: Ghanem *et al.* [3] annotated ground truth for a single dominant MF of each scene in the dataset for quantitative evaluation.
3) **3D Bounding box annotation**: Recently, Song *et al.* [44] published a new RGB-D benchmark, so-called SUN RGB-D dataset, acquired by using four different RGB-D sensors, and this benchmark involves the NYUv2 dataset as the one captured by Kinect v1 with additional useful annotations of scenes as follows: a 3D polyhedron room layout and 3D bounding boxes of scene objects (see an example in Fig. 14).

Despite the above-mentioned efforts, there has not been ground truth benchmark for multiple MFs evaluation quantitatively. Inspired by this observation and necessity of the ground truth, we propose an MMF ground truth evaluation protocol and a dataset on top of the NYUv2 dataset.

To newly annotate an MMF ground truth for the NYUv2 dataset, we leverage the above-mentioned information provided by [2], [3], [44]. The process to obtain the MMF ground truth is as follows:

1) We first separate a multiple-MF (MMF) subset from the full set of NYUv2 dataset by discarding the scenes that only have a single MF. The numbers of MFs in the scenes are provided by Straub *et al.* [2].[19] Among 1449 scenes of the full set of the NYUv2, 173 scenes are categorized as the MMF subset, 1252 scenes being single MF subset, and 24 scenes being the complement set (scenes that are not MW).
2) Given the MMF subset (173 scenes), we use the dominant MF GT provided by Ghanem *et al.* [3] as the most dominant MF of a room of each scene.
3) In addition to the most dominant MF for each scene, we annotate the other multiple-MF ground truths for the remaining dominant objects of each scene in the MMF subset. We utilize 3D bounding box annotations of scene objects in the intersection set of the SUN RGB-D with the NYUv2, where bounding boxes are aligned with 3D objects. We extract the MF information (orientation) for each object from the corresponding 3D bounding box annotation (*i.e.*, extracting surface normals of the bound boxes), and regard them as MF candidates.

---

18. http://people.csail.mit.edu/jstraub/_pages/nyu-mmf-dataset/
19. The maximum number of MFs for the NYUv2 dataset is 2.

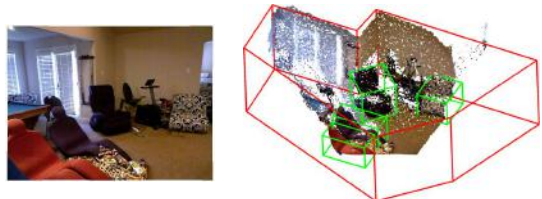

Fig. 14: Example of 3D room layout from the SUN RGB-D dataset [44] for the NYUv2 dataset [10]. *Left*: captured scene image. *Right*: 3D point clouds with the 3D room layout (red polyhedron) and orientation of scene objects (green boxes).

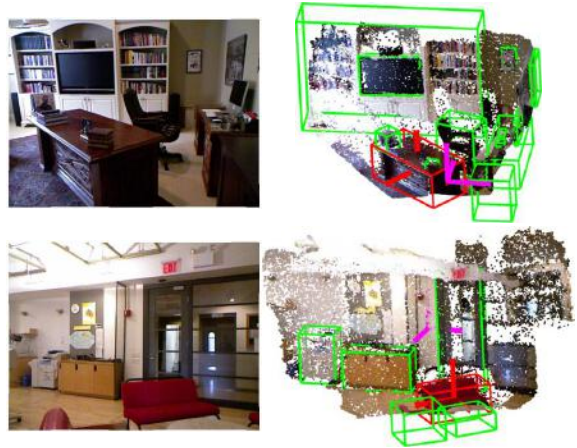

Fig. 15: Examples of manually selected MMF ground truth (2 MFs case). *Left column*: captured scene images. *Right column*: 3D point clouds with 3D bounding boxes of scene objects (MMF candidates). Green boxes indicate MMF candidates and red box is manually selected MF as the additional ground truth MFs. Red (additional MF) and magenta-colored (single dominant ground truth MF of [3]) coordinates are newly annotated MMF ground truth.

4) We remove the MF candidates that have a small angular difference ($\leq 5°$) with the most dominant MF we have extracted, and visualize the remaining MF candidates (green boxes of Fig. 15) with 3D point clouds of a scene. With the visualized interface, we manually select proper MFs (red box of Fig. 15) that satisfy the following criteria as additional MF GTs.
5) The criteria we select the MFs are as follows: 1) MFs that have the angular difference with the most dominant one larger than $5°$, and 2) MFs that are supported by at least more than $15\%$ of all valid normals (we quote this number from Straub *et al.* [2]).

We evaluate an MMF estimation algorithm by reporting each MF error among multiple MF estimates, where the error of each MF is computed as a minimum error by pairwise comparisons with all the MF GTs. Quantitative evaluation of our MMF application is reported in Table 3 of the main manuscript.

As a discussion, our dataset has a limitation at this moment. For simple and scalable annotation, we directly use the bounding box annotations given by [44], which are axis-aligned with the gravity direction and are annotated by remotely trained workers. Owing to this restriction of the bounding box, true orientations (regarded as a local MF for a specific object) of some objects are not aligned with the bounding box; for example, the red chair in Fig. 14 has a slanted surface against the gravity direction, but its annotated bounding boxes are aligned with the gravity direction. The amount of such cases is a minority of MFs. Hence, we ignore the cases at this moment, but it could be improved by manual correction. We leave this for future research.

We will share this data so that other future research can be benefited by readily benchmarking their algorithms.

## 11 ANALYSES OF VALIDITY, CONVERGENCE AND GLOBAL OPTIMALITY OF THE PROPOSED BOUNDS

### 11.1 Preliminary for Proof

We first introduce a few preliminaries for proving the theoretical guarantees of the proposed algorithm. We use the same notation defined in Table 1 in the main paper.

**Lemma 2.** *For any vector $\mathbf{v}_1$ and $\mathbf{v}_2$, $\theta(\mathbf{v}_1, \mathbf{v}_2) \leq \angle(\mathbf{v}_1, \mathbf{v}_2)$.*

*Proof.* By triangle inequality,

$$\angle(\mathbf{v}_1, \mathbf{e}_2) \leq \angle(\mathbf{v}_1, \mathbf{v}_2) + \angle(\mathbf{e}_2, \mathbf{v}_2), \text{ and}$$
$$\angle(\mathbf{e}_2, \mathbf{v}_2) \leq \angle(\mathbf{v}_1, \mathbf{v}_2) + \angle(\mathbf{v}_1, \mathbf{e}_2).$$

Rearranging the terms,

$$|\angle(\mathbf{v}_1, \mathbf{e}_2) - \angle(\mathbf{e}_2, \mathbf{v}_2)| \leq \angle(\mathbf{v}_1, \mathbf{v}_2)$$

□

The relationship between the difference along the azimuth $\phi(\cdot, \cdot)$ and $\angle(\cdot, \cdot)$ can be defined similarly. We omit the proof for this case.

**Lemma 3** (Lemma 1. in Hartley and Kahl [31]). *For any vector $\mathbf{v}$ and two rotations $\mathbf{R}$ and $\mathbf{R}'$, $\angle(\mathbf{R}\mathbf{v}, \mathbf{R}'\mathbf{v}) \leq d_\angle(\mathbf{R}, \mathbf{R}')$, where $d_\angle(\mathbf{R}, \mathbf{R}')$ is the angle lying in the range $[0, \pi]$ of the rotation $\mathbf{R}'\mathbf{R}^{-1}$.*

*Proof.* Refer to Hartley and Kahl [31]. □

**Corollary 1** (Hartley and Kahl [31]). *Given a cube $D$ and the rotation $\bar{\mathbf{R}}$ corresponding to the center of the cube, for any rotation $\mathbf{R} \in D$, we represent the radius of $r_D$ of the cube in terms of the angle metric as $r_D = \max_{\mathbf{R} \in D}(d_\angle(\bar{\mathbf{R}}, \mathbf{R}))$. Then, we have $r_D \leq \sqrt{3}\sigma$, where $\sigma$ is the half side length of the cube $D$.*

*Proof.* Refer to Hartley andd Kahl [31]. □

Now, we introduce preliminaries that are helpful for understanding mapping relationships among coordinates.

**Definition 2** (Homeomorphism). *A mapping $f : X \to Y$ between two topological spaces $(X, \tau_X)$ and $(Y, \tau_Y)$ is called a homeomorphism if it has the following properties: 1) $f$ is a bijection, 2) $f$ is continuous, and 3) $f^{-1}$ is continuous.*

**Lemma 4.** *Equirectangular projection mapping is a homeomorphism between the spherical coordinate $[0, 2\pi) \times [0, \pi]$ on $S^2$ and the equirectangular coordinates $[0, 2\pi) \times [0, \pi]$.*

*Proof.* Consider the mapping $f : [0, 2\pi) \times [0, \pi] \to [0, 2\pi) \times [0, \pi]$ defined by the equirectangular projection from a spherical coordinate $(\theta, \phi)$ into an equirectangular coordinate $(u, v)$. Since $f(\theta, \phi) = (\theta, \phi) = (u, v)$ is an identity mapping, $f$ is bijection and continuous, and $f^{-1} = f$ is continuous. Thus, $f$ is a homeomorphism. □

With homeomorphism, we will establish the sequential relationship between mappings, so that the behavior of the proposed rectangular boundary is captured by the original circular boundary on the sphere. As an object to analyze, we use the concept of the boundary set. The boundary set arises when a measurement is classified by an inlier threshold to count inlier cardinality given a rotation $\mathbf{R}$. Although boundary sets and their relationships are already described in Sec. 5.2, we specify them for general cases.

**Definition 3** (Characterization of the boundary sets). *Consider $\mathbf{R} \in SO(3)$, and an arbitrary inlier threshold $\tau' > 0$. A boundary set $\boldsymbol{\mathcal{X}}$ of inlier regions is characterized by $\mathbf{R}$ and $\tau'$. We denote this characterization of the boundary set as $\boldsymbol{\mathcal{X}}(\mathbf{R}, \tau')$. Also, we denote the closed inlier region defined by the boundary set $\boldsymbol{\mathcal{X}}(\mathbf{R}, \tau')$ as $Area(\boldsymbol{\mathcal{X}}(\mathbf{R}, \tau'))$.*

Depending on the context, we can also use the vector representation $\{\mathbf{r}_j = \mathbf{R}\mathbf{e}_j\}_{j=1}^6$ instead of $\mathbf{R} \in SO(3)$ in Definition 3, *e.g.*, $\boldsymbol{\mathcal{X}}(\{\mathbf{r}_j = \mathbf{R}\mathbf{e}_j\}_{j=1}^6, \tau')$. Additionally, we say that $\boldsymbol{\mathcal{X}}$ is equal to $\boldsymbol{\mathcal{X}}'$ iff $Area(\boldsymbol{\mathcal{X}}) = Area(\boldsymbol{\mathcal{X}}')$.

**Lemma 5** (Uniqueness of characterization). *A boundary set $\boldsymbol{\mathcal{X}}(\{\mathbf{r}_j = \mathbf{R}\mathbf{e}_j\}_{j=1}^6, \tau')$ is uniquely characterized by $(\mathbf{R}, \tau')$.*

*Proof.* Consider $\boldsymbol{\mathcal{X}}_\mathbf{R}$ characterized by $(\{\mathbf{r}_j\}_{j=1}^6, \tau')$, and assume that there exists another $(\{\mathbf{r}^\#_j\}_{j=1}^6, \tau^\#)$ characterizing $\boldsymbol{\mathcal{X}}_\mathbf{R}$. This implies $Area(\boldsymbol{\mathcal{X}}(\{\mathbf{r}_j\}_{j=1}^6, \tau')) = Area(\boldsymbol{\mathcal{X}}(\{\mathbf{r}^\#_j\}_{j=1}^6, \tau^\#))$.

By definition, the direction vector passing through the center of mass of each boundary $\mathcal{X}_j$ is $\mathbf{r}_j$, and the center of mass is unique, *i.e.*, $\mathbf{r}_j = \mathbf{r}^\#_j$. Also, $\tau'$ ($\tau^\#$) is a geodesic radius between a boundary and $\mathbf{r}_j$ ($\mathbf{r}^\#_j$) which specify $Area(\boldsymbol{\mathcal{X}})$ as shown in Fig. 16-[Left]. Thus, given $\mathbf{r}_j = \mathbf{r}^\#_j$, in order to be $Area(\boldsymbol{\mathcal{X}}(\{\mathbf{r}_j\}_{j=1}^6, \tau')) = Area(\boldsymbol{\mathcal{X}}(\{\mathbf{r}^\#_j\}_{j=1}^6, \tau^\#))$, $\tau^\#$ should be $\tau'$. This contradicts the assumption; hence, the characterization is unique. □

We have three types of boundary sets: the original boundary set $\boldsymbol{\mathcal{X}}$ on $S^2$ (3D spherical measurement or 3D-EGI domain) in Fig. 2a, the curved boundary set $\hat{\boldsymbol{\mathcal{X}}} = \{\hat{\mathcal{X}}_j\}_{j=1}^6$ on the 2D-EGI domain in Fig. 2b, and the rectangular boundary set, denoted as $\boldsymbol{\mathcal{R}} = \{\mathcal{R}_j\}_{j=1}^6$, on the 2D-EGI domain in Fig. 2d.

Note that each boundary set is a set of boundaries determined by $\{\mathbf{r}_j\}_{j=1}^6$ and a given threshold $\tau'$.[20] Namely, three types of boundary sets are characterized in the same manner with Definition 3. Since we have not shown any property of the characterizations for $\hat{\boldsymbol{\mathcal{X}}}$ and $\boldsymbol{\mathcal{R}}$, we now introduce properties of their characterization and relationships.

Specifically, a boundary $\mathcal{X}_j$ is a small circle on $S^2$ manifold of which center point is passed through by $\mathbf{r}_j$, *i.e.* a rotation axis, and of which geodesic radius is $\tau'$. Following the procedure to get a curved boundary set and a rectangular one described in Sec. 5, they are also uniquely characterized by $(\mathbf{R}, \tau')$ as the equirectangular projection is homeomorphic. Namely, once a curved and a rectangular boundary set is derived from a boundary set $\boldsymbol{\mathcal{X}}(\mathbf{R}, \tau')$, then the three types of boundary sets share the same characterization parameter $(\mathbf{R}, \tau')$, as illustrated in Fig. 16.

---

20. For the boundary sets defined on the 2D-EGI domain, we represent the center point of a boundary (*i.e.*, rotation axis) with azimuth and elevation angles as $\mathbf{r}_j = [r^j_\theta, r^j_\phi]$.

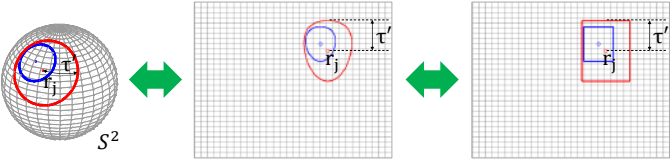

Fig. 16: Illustration of homeomorphism between the spherical coordinate and the equirectangular coordinate. Green double-sided arrow means bijective relationship. Three types of boundary sets share the same characterization parameter ($\{\mathbf{r}_j\}_{j=1}^6, \tau'$). For only visualization purpose, we draw each boundary, $\mathcal{X}_j$, $\hat{\mathcal{X}}_j$, and $\mathcal{R}_j$ associated with $\mathbf{r}_j$. Notice that the red boundary is in contact with the blue boundary, and this relationship also holds for all the domains, and likewise, the inclusion relationship between the red and blue boundaries also holds for all the domains. As another point, the geodesic radius $\tau'$ of the small red circle on $S^2$ is preserved as the half height on the 2D-EGI domain, as the equirectangular projection does not distort along the elevation direction.

Conversely, given any boundary set $\mathcal{X}$, we can inversely find at least two center points out of six boundaries in the set and can calculate the inlier threshold $\tau'$ by the relationship between the center and boundary (see Fig. 16) on a domain (3D or 2D-EGI), *e.g.*, the half height of a boundary in 2D-EGI.[21] Then, by using the orthogonality among $\{\mathbf{r}_j\}$, we can uniquely determine the shared characterization parameter ($\{\mathbf{r}_j\}, \tau'$).

From this, we have the following relationship between boundaries.

**Lemma 6.** *Let $\mathbf{R} \in SO(3)$ be an arbitrary rotation and $\tau' > 0$ be an arbitrary constant. Consider the proposed mapping between a boundary set $\mathcal{X}(\mathbf{R}, \tau')$ on $S^2$, and its corresponding rectangular boundary set $\mathcal{R}(\mathbf{R}, \tau')$ on a continuous 2D-EGI domain. The mapping is bijective (one-to-one correspondence).*

*Proof.* We prove this by showing the bijectivity between $\mathcal{X}$ and ($\{\mathbf{r}_j\}, \tau'$), $\mathcal{X}$ and $\hat{\mathcal{X}}$, and finally, $\hat{\mathcal{X}}_j$ and $\mathcal{R}$.

A. **Bijective mapping between $\mathcal{X}$ and ($\{\mathbf{r}_j\}, \tau'$):** By Lemma 5, a boundary set $\mathcal{X}$ on $S^2$ is uniquely characterized as ($\{\mathbf{r}_j\}, \tau'$). This characterization can be viewed as an injective mapping, *i.e.*, one-to-one mapping. Its inverse can be done by finding directions $\{\mathbf{r}_j\}$ (which is a set of direction vectors to center points of each boundary $\mathcal{X}_j$) and $\tau'$ (which is the geodesic radius between a boundary $\mathcal{X}_j$ and $\mathbf{r}_j$), *i.e.*, bijective. Its illustrations refer to Figs. 2a and 16-[Left].

B. **Bijectivity between $\mathcal{X}$ and $\hat{\mathcal{X}}$:** Point elements in $\mathcal{X}_j$ have a one-to-one correspondence to point elements in the corresponding boundary $\hat{\mathcal{X}}_j$ on 2D-EGI by the equirectangular projection mapping, which is a bijective mapping. $\mathbf{r}_j$ is also bijectively mapped onto 2D-EGI. Thus, $\hat{\mathcal{X}}$ can be characterized by ($\{\mathbf{r}_j\}, \tau'$), where $\tau'$ can be obtained from the half height of a boundary of $\{\mathcal{X}_j\}$ in 2D-EGI. This bijective mapping between $\mathcal{X}_j$ and $\hat{\mathcal{X}}_j$ is illustrated in Figs. 2a, 2b and 16-[Left, Middle].

C. **Bijectivity between $\hat{\mathcal{X}}_j$ and $\mathcal{R}$:** From the definition described in Sec. 5.2, the rectangular boundary set $\mathcal{R}$ is uniquely determined as tightest and axis-aligned rectangles enclosing (circumscribing) each boundary $\hat{\mathcal{X}}_j$, which is an injective mapping, and is also characterized as ($\{\mathbf{r}_j\}, \tau'$). Now, we show that the mapping between $\hat{\mathcal{X}}$ and $\mathcal{R}$ is surjective. We restrict the set of rectangles on 2D-EGI by the above characterization and can regard the set of restricted rectangles as a codomain $Y$. Let $X$ be the set of all $\hat{\mathcal{X}}$, *i.e.*, domain. Once a boundary set $\mathcal{R}$ is characterized as ($\{\mathbf{r}_j\}, \tau'$), then $\hat{\mathcal{X}}$ can be characterized with the same parameters. This relationship exactly follows the definition of surjective mapping ($\forall y \in Y, \exists x \in X$ such that $f(x) = y$). Thus, the mapping is bijective. The relationship is illustrated in Figs. 2b, 2d, 4b and 16-[Middle, Right].

---

21. The height has no distortion by the mapping between $S^2$ and 2D-EGI. One can understand this fact by the Tissot's indicatrix of the equirectangular projection.

Since the composition of bijective mappings is bijective, the composition from A to C is also bijective, *i.e.*, one-to-one correspondence between $\mathcal{X}$ and $\mathcal{R}$. This completes the proof. □

**Remark 1.** *Lemmas 4 and 6 show the topological relationship (homeomorphism) between the boundary set $\mathcal{X}$ on the spherical domain $S^2$ and the proposed rectangular boundary set $\mathcal{R}$ on the 2D-EGI. For instance, let $\mathcal{R}$ ($\mathcal{R}'$) be induced by $\mathcal{X}$ ($\mathcal{X}'$), then 1) when $Area(\mathcal{X})$ includes another $Area(\mathcal{X}')$ on the spherical domain, then $\mathcal{R}$ and $\mathcal{R}'$ have the same inclusion relationship on 2D-EGI, and 2) when $\mathcal{X}$ converges to $\mathcal{X}'$ on $S^2$, then $\mathcal{R}$ also converges to $\mathcal{R}'$ on 2D-EGI.*

These allow us to analyze the property of the rectangular boundary set by analyzing $\mathcal{X}$ on $S^2$ instead. Thus, we can prove related statements to the rectangular bounds indirectly by $\mathcal{X}$ yet exactly instead of directly analyzing the rectangular boundary set.

We also introduce a probabilistic interpretation of 2D-EGI as a simple analysis tool. Since 2D-EGI is essentially similar to a density function (continuous) or a 2D histogram (discrete), by normalizing values of entries by integration over all the domain, we can interpret it as a probability density function $P(\cdot)$. This is seamless in that the cardinality maximization problems we deal with are scale-invariant.

### 11.2 Proof of Lemma 1a

*Proof.* Let $\mathbf{R}^*$ be the optimal rotation for a cube $D$ providing $c_D^*$. Since $c_D^*$ is the optimum for all $\mathbf{R} \in D$, there is no rotation that provides larger cardinality than $c_D^*$ in $D$. That is, the inlier cardinality of any rotation $\mathbf{R} \in D$ is smaller or equal to $c_D^*$, *i.e.*, $L_R \leq c_D^*$.

The validity of upper bound $U_R$ can be proved using the topological inclusion relationship induced by Lemmas 4 and 6 (homeomorphism).

Corollary 1 implies that, for a center rotation $\bar{\mathbf{R}}$ of $D$ and a threshold $\tau + \sqrt{3}\sigma$, the boundary set $\mathcal{X}(\bar{\mathbf{R}}, \tau + \sqrt{3}\sigma)$ on $S^2$ includes any boundary set $\mathcal{X}(\mathbf{R}, \tau)$ for $\mathbf{R} \in D$ (This is already proved in Bazin *et al.* [22]). This is because any inlier of ($\mathbf{R}, \tau$) is also an inlier of ($\bar{\mathbf{R}}, \tau + \sqrt{3}\sigma$). We call this an *upper boundary inclusion*. This inclusion relationship is illustrated in Fig. 16-[Left].

Note that this inclusion relationship applies only to the topological relationship between boundaries, not to the upper bound value of $U_B$.

Since the *upper boundary inclusion* is also satisfied for the rectangular boundary case by virtue of the homeomorphism (Lemma 6), $\mathcal{R}_\mathbf{U}$ by Eq. (7) for a center rotation $\bar{\mathbf{R}}$ of $D$ includes any $\mathcal{R}(\mathbf{R}, \tau)$ for $\mathbf{R} \in D$. Thus, $\mathcal{R}_\mathbf{U}$ also includes $\mathcal{R}^* = \mathcal{R}(\mathbf{R}^*, \tau)$, *i.e.*, $Area(\mathcal{R}^*) \subseteq Area(\mathcal{R}_\mathbf{U})$. Since $P(A) \leq P(B)$ if $A \subseteq B$, $P(Area(\mathcal{R}^*)) \leq P(Area(\mathcal{R}_\mathbf{U}))$. This implies $c_D^* \leq U_R$, which completes the proof. □

### 11.3 Proof of Lemma 1b

*Proof.* We now show the convergence of the bound stated as the following:

$$\forall \epsilon \geq 0, k \in \{1, 2, ...\}, \text{ and } D \subseteq D_0, \exists \delta \geq 0 \text{ such that}$$
$$\text{size}(D) \leq \delta \Rightarrow U_R^{D,k} - L_R^{D,k} \leq \epsilon, \quad (11)$$

where $D_0$ denotes the initial cube enclosing the solution search space and $U_R^{D,k}$ denotes the upper bound obtained by the system in Eq. (7) at iteration $k$ with the half side length $\sigma_k$ of the cube $D$ (analogously $L_R^{D,k}$ denotes the lower bound obtained by Eq. (6)).

First, for large $k$, the divided cubes at the $k$-iteration must be small, *i.e.* $\text{size}(D) \leq \frac{\sqrt{3}\sigma_0}{2^k}$, by Corollary 1. As $k \to \infty$, it converges to zero. This indicates that $\delta \geq 0$ exists for any $k$.

We now show the right bound over all $k$. To easily find the $\epsilon$-bound of the rightmost term in Eq. (11), we use Lemma 6 again. Analogous to the proof of Lemma 1a, when the gap between $\mathcal{X}_\mathbf{U}$ and $\mathcal{X}_\mathbf{L}$ shrinks as $\sigma_k \to 0$ by subdivision, the corresponding gap between $\mathcal{R}_\mathbf{U}$ and $\mathcal{R}_\mathbf{L}$ also shrinks (by homeomorphism).

The inlier boundaries on the spherical domain are determined from the systems in Eqs. (3) and (4)[22] as $\mathcal{X}_\mathbf{U}$ of $(\{\bar{\mathbf{r}}_j\}, \tau + \sqrt{3}\sigma_k)$ for the upper bound and $\mathcal{X}_\mathbf{L}$ of $(\{\bar{\mathbf{r}}_j\}, \tau)$ for the lower bound, where $\bar{\mathbf{R}}$ corresponds to the center of the cube $D$. Note that $\mathcal{X}_\mathbf{L}$ is invariant to $k$, while $\mathcal{X}_\mathbf{U}$ depends on $k$; hence we denote $\mathcal{X}_{\mathbf{U}^k}$ according to $k$. The gap of the area between $\mathcal{X}_{\mathbf{U}^k}$ and $\mathcal{X}_\mathbf{L}$ is proportional to $(\tau + \sqrt{3}\sigma_k) - \tau = \sqrt{3}\sigma_k = \frac{\sqrt{3}\sigma_0}{2^k}$ (by the subdivision rule of the algorithm, we have $\sigma_k = \frac{\sigma_0}{2^k}$). It means that the area difference between the bound regions shrinks according to increasing $k$. This relationship is illustrated in Fig. 17a.

We only borrow this boundary shrink property of $\mathcal{X}$, in that when there exists $\mathcal{X}$ converging to $\mathcal{X}'$ on $S^2$, then so is $\mathcal{R}$ corresponding to $\mathcal{X}$ converging to $\mathcal{R}'$ corresponding to $\mathcal{X}'$ on 2D-EGI. To move forward, we need the following lemma.

**Lemma 7** (Continuity of Probability [63]). *Suppose $A_1, A_2, \cdots$ is a sequence of events.*

*1) If $A_1 \subset A_2 \subset \cdots$, then $\lim_{j \to \infty} P(A_j) = P(\cup_{i=1}^\infty A_i)$.*
*2) If $A_1 \supset A_2 \supset \cdots$, then $\lim_{j \to \infty} P(A_j) = P(\cap_{i=1}^\infty A_i)$.*

By the concept of *upper boundary inclusion* in the proof of Lemma 1a, we have $Area(\mathcal{R}_{\mathbf{U}^k}) \supseteq Area(\mathcal{R}_\mathbf{L})$. Denote $A_k = Area(\mathcal{R}_{\mathbf{U}^k})$ such that $A_0 \supset A_1 \supset \cdots \supset A_\infty = Area(\mathcal{R}_\mathbf{L})$. Then, by the second statement of Lemma 7, we have $\lim_{k \to \infty} P(A_k) = P(Area(\mathcal{R}_\mathbf{L}))$. With a proper scaling, by the boundness of probability space, we have $U_R - L_R \leq \epsilon$. □

### 11.4 Proof of Proposition 2.

*Proof.* First, suppose a case of $s \to \infty$, *i.e.*, the measurement space (EGI) is continuous, on which the classic convergence result of BnB is given. The BnB is guaranteed to converge to the globally optimal solution when the lower and upper bounds are valid [17], [34]. Hence, once all the bounds $L_R$ and $U_R$ are valid, the proposed BnB framework provides global optimality. We have already proven that $L_R$ and $U_R$ are valid bounds by Lemma 1. By the results in [17], [34], the BnB algorithm with the proposed bounds in Eqs. (6) and (7) is guaranteed to converge to a globally optimal solution once it converges to a point estimate.

---

22. Notice again that we do not use the values of lower and upper bounds, but use the topological relationship of boundaries.

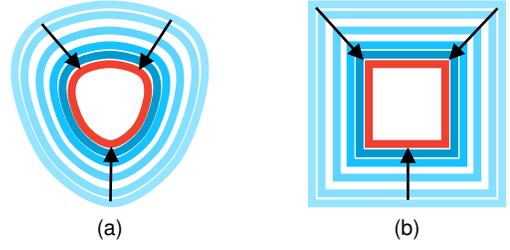

Fig. 17: Illustration of convergence. (a) The curved boundaries $\hat{\mathcal{X}}$. (b) The rectangular boundaries $\mathcal{R}$.

Consider $s$ is finite, *i.e.*, EGI is a discretized domain as described in this work. On the discrete domain, there is a certain sub-division level $K$ that no longer reduces the gap between the subsequent upper bound values, *i.e.*, $\forall k \geq K, \exists \epsilon' > 0$ such that $|U_R^{k+1} - U_R^k| \geq \epsilon'$ due to the resolution of EGI. We can choose the stop criterion $\xi$ by this fact as the difference of the thresholds between the subsequent upper bound systems.

$$\xi_k = (\tau_{\text{el}} + \sqrt{3}\sigma_k) - (\tau_{\text{el}} + \sqrt{3}\sigma_{k+1}) = \sqrt{3}(\sigma_k - \sigma_{k+1}), \quad (12)$$

where $k$ is the number of iteration (equal to the level of subdivision). For simplicity, we prove the convergence behavior only for the elevation axis in this proof, because we can separately consider the elevation and azimuth axis thresholds (we can stop the algorithm when both or either stop criteria are satisfied). However, the case of the azimuth axis can be proved in the same way.

Since we have $\sigma_k = \frac{\sigma_0}{2^k}$ by the sub-division rule, where $\sigma_0$ denotes the half-side length of the initial cube, we can rewrite $\xi_k = \frac{\sqrt{3}\sigma_k}{2}$. Because of the resolution limit by $1/s°$, when $\xi_k \leq 1/s°$, more branching does not improve the accuracy. Thus, we have the following criterion:

$$\xi_k = \frac{\sqrt{3}\sigma_k}{2} \leq \frac{1}{s}°. \quad (13)$$

Once the algorithm converges to or is terminated by this stop criterion at the iteration $k$ with $\mathbf{R}'$ near the global optimal $\mathbf{R}^*$, for any vector $\mathbf{v}$, we have

$$\begin{aligned}
\theta(\mathbf{R}'\mathbf{v}, \mathbf{R}^*\mathbf{v}) &\leq \angle(\mathbf{R}'\mathbf{v}, \mathbf{R}^*\mathbf{v}) &&\text{(by Lemma 2)} \\
&\leq d_\angle(\mathbf{R}', \mathbf{R}^*) &&\text{(by Lemma 3)} \\
&\leq r_D \leq \sqrt{3}\sigma_k &&\text{(by Corollary 1)} \\
&\leq 2/s°. &&\text{(by Eq. (13))}
\end{aligned} \quad (14)$$

The proof for the azimuth criterion is analogous to above. We can choose $\epsilon$ to be $2/s°$, where $s$ is the parameter determined by the user for the EGI resolution. Consequently, there exists a constant $\epsilon$ that guarantees $\epsilon$-optimality. The proof is completed. □